\documentclass[acmtog]{acmart}
\usepackage[capitalize,nameinlink]{cleveref}
\usepackage{enumitem}
\usepackage{subcaption}
\acmSubmissionID{1103}

\usepackage{pifont}

\DeclareGraphicsExtensions{.png,.jpg,.pdf,.ai,.psd}
\newcommand{\mysection}[2]{%
    \section{#1}%
    \label{sec:#2}%
}

\newcommand{\mysubsection}[2]{%
    \subsection{#1}%
    \label{sec:#2}%
}

\definecolor{colorA}{HTML}{4285f4}
\definecolor{colorB}{HTML}{ea4335}
\definecolor{colorC}{HTML}{fbbc04}
\definecolor{colorD}{HTML}{34a853}
\definecolor{colorE}{HTML}{ff6d01}
\definecolor{colorF}{HTML}{46bdc6}
\definecolor{colorG}{HTML}{000000}
\definecolor{colorH}{HTML}{777777}
\definecolor{colorI}{HTML}{bdd6ff}
\definecolor{colorJ}{HTML}{6a9e6f}

\providecommand{\checkmark}{\ding{51}}

\providecommand{\xmark}{\scalebox{0.85}{\ding{53}}}

\newcounter{fact}

\definecolor{siggreen}{RGB}{0,140,90}
\renewcommand{\checkmark}{\textcolor{siggreen}{\ding{51}}}
\renewcommand{\xmark}{\textcolor{red}{\ding{55}}}

\copyrightyear{2026}
\acmYear{2026}
\acmConference[SIGGRAPH Conference Papers '26]{Special Interest Group on Computer Graphics and Interactive Techniques Conference Conference Papers}{July 19--23, 2026}{Los Angeles, CA, USA}
\acmBooktitle{Special Interest Group on Computer Graphics and Interactive Techniques Conference Conference Papers (SIGGRAPH Conference Papers '26), July 19--23, 2026, Los Angeles, CA, USA}
\acmDOI{10.1145/3799902.3811186}
\acmISBN{979-8-4007-2554-8/2026/07}

\begin{document}
\begin{teaserfigure}
    \centering
    \includegraphics{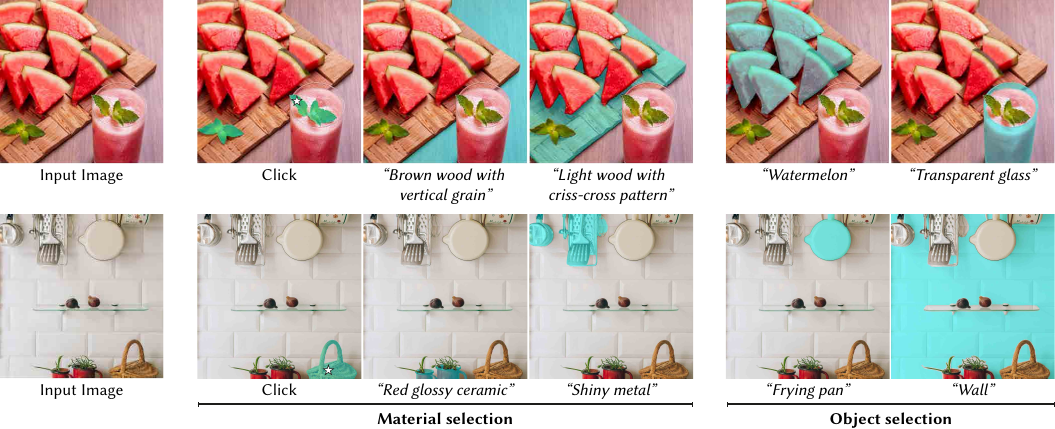}
    \caption{
    Our method, MAOAM, enables click- and text-based selection of both materials and objects via single model with a unified interface. Given an input image, users can interact with the selection model via clicks (second column, denoted as overlaid white star) or text queries (third column onwards). 
    }
    \label{fig:teaser}%
\end{teaserfigure}

\newcommand{\bftab}[1]{{\fontseries{b}\selectfont#1}}

\newcommand{\dataset}[1]{\textsc{#1}}
\newcommand{\best}[1]{\textbf{#1}}

\title{MAOAM: Unified Object and Material Selection with Vision-Language Models}

\author{Jaden Park}
\affiliation{%
  \institution{University of Wisconsin-Madison}
  \country{USA}
}
\orcid{0000-0003-1901-4157}
\email{jadenpark@cs.wisc.edu}

\author{Valentin Deschaintre}
\affiliation{%
    \institution{Adobe Research}
    \country{United Kingdom}
}
\email{deschain@adobe.com}
\orcid{0000-0002-6219-3747}

\author{Jason Kuen}
\affiliation{%
    \institution{Adobe Research}
    \country{USA}
}
\email{kuen@adobe.com}
\orcid{0000-0001-5099-8145}

\author{Kangning Liu}
\affiliation{%
    \institution{Adobe Research}
    \country{USA}
}
\email{kangningl@adobe.com}
\orcid{0000-0002-0187-4602}

\author{Iliyan Georgiev}
\affiliation{%
    \institution{Adobe Research}
    \country{United Kingdom}
}
\email{igeorgiev@adobe.com}
\orcid{0000-0002-9655-2138}

\author{Krishna Kumar Singh}
\affiliation{%
    \institution{Adobe Research}
    \country{USA}
}
\email{krishsin@adobe.com}
\orcid{0000-0002-8066-6835}

\author{Yong Jae Lee}
\affiliation{%
    \institution{Adobe Research}
    \country{USA}
}
\email{yongl@adobe.com}
\orcid{0000-0001-9863-1270}

\author{Michael Fischer}
\affiliation{%
    \institution{Adobe Research}
    \country{United Kingdom}
}
\email{mifischer@adobe.com}
\orcid{0000-0002-2610-4831}

\renewcommand{\shortauthors}{Park et al.}

\begin{abstract}

Selection is a core operation in interactive image editing, enabling tasks such as composition or manipulation. 
To be practically useful, a user should be able to specify and disambiguate the desired selection region through either text- or click-based interactions, and the system should support selecting not only objects but also other criteria, such as materials.  Material-based selection can be particularly valuable for tasks like re-texturing surfaces or consistently editing all instances of a specific material in a scene. However, existing vision–language-model (VLM) based selection methods are largely object-centric and typically support only a single interaction modality, limiting their applicability in real editing workflows.

In this work, we thus present \underline{M}ask \underline{A}ny \underline{O}bject \underline{A}nd \underline{M}aterial (MAOAM), a unified selection framework that enables precise object- and material-level selection across both text- and click-based interactions. MAOAM leverages a VLM with a segmentation head to produce pixel-accurate masks from user prompts: the VLM interprets the user's selection intent --- object- or material-level --- and encodes visual entities, attributes, and spatial relations, while the segmentation head decodes the VLM's output token into a mask.

A key challenge is that material selection datasets with text annotations are unavailable. We therefore propose a scalable data generation pipeline: we collect real and synthetic images with material masks, then leverage VLMs to generate material descriptions with rich visual-semantic information. Using the generated data, we train MAOAM with a multi-task objective over click- and text-based selection, along with an auxiliary VQA task derived from the material descriptions to facilitate deeper material understanding.

Despite being trained with uni-modal prompts, our model exhibits an emergent improvement in selection quality when combining text and clicks at inference time, enabling more flexible image editing workflows. Experiments demonstrate accurate and coherent selections across diverse objects, materials, and interaction scenarios, highlighting robustness in practice.

\end{abstract}


\maketitle

\mysection{Introduction}{introduction}

Selection --- producing pixel-accurate segmentation masks under user-specified criteria --- is a fundamental operation in interactive image editing, enabling downstream tasks such as compositing, relighting, and appearance manipulation.
In practice, users vary both in \emph{how} they interact with the system --- e.g., clicks, which are precise and local, or text prompts, which can describe the desired region using complex visual qualities or spatial relations --- and in \emph{what} they would like to select (e.g., objects or materials). Importantly, a single interaction modality can be insufficient to fully disambiguate the intended selection criterion. For example, consider \textit{selecting all ceramic plates} in a kitchen with both ceramic and plastic plates, and a ceramic pot. Uni-modal queries such as \textit{``select the plates''} would include the plastic ones, while \textit{``select the ceramic''} would also include the pot. Clicking each ceramic plate quickly becomes impractical when many instances are present. The most natural and efficient query \textit{``select all ceramic plates''} requires joint reasoning over object and material within a single model. Hence, an ideal interactive selection model should support multiple interaction modalities (clicks, text, or both) and diverse selection criteria beyond objects alone.

Recent segmentation models --- ranging from the Segment Anything Model (SAM) series~\citep{kirillov2023segany, ravi2024sam2} to VLM-based approaches~\citep{lai2024lisa,rasheed2024glamm,zhang2024evf,yuan2025sa2va} --- mostly support a single interaction modality (clicks or text) and exclusively focus on object-level selection. 
Material selection, in contrast, has different semantics and structure: a single material may span multiple objects (e.g., metal across fixtures) or appear as disjoint sub-regions within an object (e.g., the wooden legs of a chair). This capability is particularly valuable for tasks like re-texturing surfaces or consistently editing all instances of a specific material in a scene.
Prior works on material selection~\cite{sharma2023materialistic, guerrero2025matselection, fischer2026sama} only allow click-based interactions, limited to local spatial cues and thus cannot explicitly express global or relative semantic criteria (e.g., \emph{all} glossy metal; the fabric \emph{in the back}) or disambiguate whether the intended criterion is an object or a material. %

We propose a unified selection framework supporting both object- and material-level segmentation across both text- and click-based interactions. Given a natural-language prompt or a click, our system produces a mask consistent with the user's selection criterion. We leverage VLMs to handle both modalities — text lets users describe complex visual semantics and attributes, while clicks specify precise locations. We build on prior VLM-based approaches~\citep{lai2024lisa,rasheed2024glamm,yuan2025sa2va} and extend it to unified object and material segmentation. Conditioned on the user prompt, the VLM processes the image and emits a segmentation token, which a segmentation head decodes into the mask, encoding the visual and spatial information needed for the target selection. To strengthen material understanding, we utilize a multi-task objective combining segmentation with a VQA-based reasoning task.

\begin{table}[t]
\centering
\caption{Supported criteria and input modalities across methods. Only MAOAM supports both object and material selection under both click- and text-based interaction.}
\label{tab:input_modalities}
\resizebox{\columnwidth}{!}{%
\begin{tabular}{lccccccc}
\toprule
 & LISA & GLaMM & Sa2VA & SAM3 & SAMa & Materialistic & \textbf{MAOAM} \\
\midrule
Objects        & \checkmark & \checkmark & \checkmark & \checkmark & \xmark & \xmark & \checkmark \\
Materials      & \xmark     & \xmark     & \xmark     & \xmark     & \checkmark & \checkmark & \checkmark \\
Text Input   & \checkmark & \checkmark & \checkmark & \checkmark & \xmark & \xmark & \checkmark \\
Click Input    & \xmark     & \xmark     & \xmark     & \checkmark & \checkmark & \checkmark & \checkmark \\
\bottomrule
\end{tabular}}
\end{table}

A central challenge in training our unified selection model is the lack of text-annotated material datasets. Existing segmentation datasets are object-centric and do not generalize to materials, as one object may comprise multiple materials, and multiple objects may consist of the same material. 
In addition, the material assignments in existing datasets \cite{guerrero2025matselection, fischer2026sama} are semantically inconsistent (e.g., a plate made of fabric) which limits their use in teaching a model about real-world material appearance. 

To address these limitations, we first collect real-world and rendered synthetic sets of images with highly precise material masks. We then propose a scalable data generation pipeline that leverages advanced VLMs to densely annotate materials with text descriptions rich in visual semantics and spatial information. With the generated descriptions, we carefully formulate VQA questions to encourage fine-grained understanding of material qualities in the text space, and jointly train our model along with the selection task.

We show our data generation and training strategy improves material selection and understanding while maintaining competitive object selection performance to several strong baselines. Notably, due to the diversity in our material description data, the model learns to handle text input of varying complexity (e.g., from \textit{``select the wood''} to \textit{``select the red-brown wood with vertical grains and gentle weathering''}), demonstrating semantic grounding of material descriptions in the image space. %
Although the model is trained with uni-modal data consisting of \emph{either} click- or text-based prompts, we observe an emergent improvement in selection when combining text and clicks during inference, enabling more flexible image editing workflows (see \cref{fig:teaser}). 
Our contributions are as follows:
\begin{itemize}[leftmargin=*,topsep=0pt,labelsep=0.5em]
    \item We propose a unified model that produces object or material selection masks from both click- and text-based interactions.
    \item We collect selection data with material-level annotations and design a scalable VLM-based data generation pipeline that generates semantically rich and grounded descriptions of materials.
    \item We empirically validate that our text-data generation pipeline leads to generalization to diverse user interaction patterns, enabling flexible behavior in real editing workflows.
\end{itemize}

\noindent We release our model and test code alongside evaluation data to facilitate further research in this direction \href{https://github.com/adobe-research/obj-and-mat-selection}{here}.%

\begin{figure*}
    \centering    \includegraphics{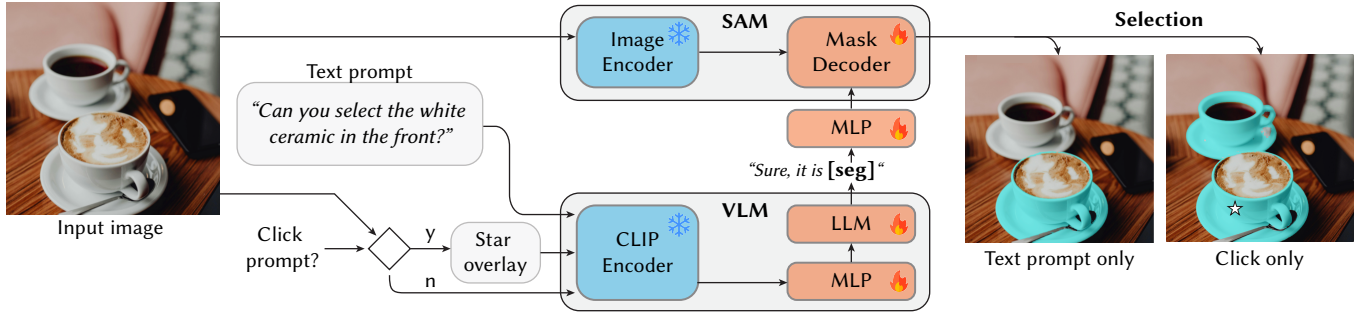}
    \caption{MAOAM architecture overview. Given an input image, MAOAM takes a task prompt specifying the selection criteria (i.e. objects or materials) alongside a user prompt to specify the desired selection in click or text. If a click is provided, stars are overlaid onto the image as visual cues. The VLM's CLIP-encoder and projection layer encode the image features into the language-space. The LLM processes the features and produces a segmentation-token, which is projected via another MLP as a prompt for the mask decoder.
    We train our model with a multi-task objective on click- and text-based selection alongside VQA, where blue and red denote frozen and trainable parameters, respectively. In the above example, notice how with click-only, all areas made of white ceramic are selected, while with the text prompt, only the front object with white ceramic is selected.%
    }
    \label{fig:main}%
\end{figure*}

\mysection{Previous Work}{previous_work}

\paragraph{Segmentation and Selection.} %
Image segmentation has long been a central problem in computer vision (for a survey, see~\cite{segmentation_survey}). Recent advances 
include bipartite matching with object queries~\cite{carion2020detr}, unified decoder-only formulations for panoptic, semantic, and instance segmentation~\cite{cheng2021maskformer}, and multi-scale feature processing for improved accuracy~\cite{Cheng_2022_mask2former}.
Beyond automatic segmentation, interactive selection has gained significant attention, particularly with the introduction of SAM~\cite{kirillov2023sam}, which enables user-guided mask generation via clicks, boxes, or points. 
Subsequent works have improved mask quality~\cite{ke2023samhq} and extended the framework to video~\cite{ravi2024sam2} and text \cite{carion2025sam3segmentconcepts}.

However, these methods are fundamentally object-centric. In concurrent work with SAM, Materialistic~\cite{sharma2023materialistic} introduced selection based on material similarity, allowing joint selection of image regions sharing the same material. 
Follow-up work further improved selection granularity~\cite{guerrero2025matselection} and extended material-based selection to video and 3D~\cite{fischer2026sama}.
Unlike prior approaches that separate object and material selection, we propose a single, unified model that supports both object- and material-level selection with text and/or click-based interactions, enabling more flexible and expressive user interaction.

\paragraph{VLM-based segmentation.} %
Recent work has augmented VLMs and multi-modal LLMs (MLLMs) with pixel-level outputs to enable grounded image understanding and segmentation from natural language. Early generalist models like X-Decoder~\cite{zou2022xdecoder} and SEEM~\cite{zou2023seem} established unified interfaces bridging pixel-level masks and vision–language semantics for segmentation tasks. LISA~\cite{lai2024lisa} introduced reasoning segmentation for implicit, knowledge-intensive queries, while GLaMM~\cite{rasheed2024glamm} advanced dense grounding by generating natural language responses intertwined with segmentation masks. Subsequent work has expanded language-guided segmentation along multiple axes: multi-target referring and explicit rejection handling (GSVA~\cite{xia2024gsva}), unified training across heterogeneous tasks (PSALM~\cite{zhang2025psalm}), efficient vision–language feature fusion (EVF-SAM~\cite{zhang2024evf}), and reasoning-centric approaches using chain-of-thought guidance (ThinkFirst~\cite{kao2025think}) or reinforcement learning (Seg-Zero~\cite{liu2025segzero}). In contrast to these object-centric frameworks, our work is the first unified VLM-driven approach to support both object- and material-level selection across text and click inputs.

\paragraph{Visual Grounding} Visual grounding studies the alignment of linguistic concepts and image regions, with the goal of localizing where a model associates words or phrases within the visual domain. Early approaches 
typically predicted bounding boxes or coarse spatial regions corresponding to textual queries~\cite{plummer2015flickr30k,yu2018mattnet}, while 
more recent work analyzes cross-attention activations between vision and language ~\cite{kang2025your}.

Grounding provides valuable model interpretability, but is not designed to produce editing-quality masks, particularly for fine-grained structures or material boundaries.  
While our method implicitly employs grounding-based localization, our goal is high-precision, interactive selection
by combining VLM-based semantic understanding with explicit user input to support accurate object- and material-level selection suitable for image editing workflows.

\mysection{Method}{method}
We describe our approach to training a unified selection model supporting click- and text-based interactions for both object- and material-level selection.
\cref{fig:main} shows an overview of our method. 

\mysubsection{Architecture}{arch}

We build upon state-of-the-art VLM-based object segmentation architectures~\cite{lai2024lisa,rasheed2024glamm,yuan2025sa2va}, which couple a vision-language model (VLM) that processes the input image and selection prompt with a SAM-based mask decoder~\cite{kirillov2023sam}, and extend them to support unified object and material segmentation. Many such systems further incorporate additional modules (e.g., 4-level region encoders or localized feature extractors) to inject spatially focused visual features directly into the decoder. In contrast, inspired by recent work that provides explicit visual cues to the VLM \cite{cai2023vipllava}, we provide the VLM with the full input image either (i) with a star overlay indicating the click location or (ii) paired with a referring text prompt. This allows us to preserve a clean, unified input interface across different architectures. The visual overlay is also required to ground the VLM in the click location for the VQA task (\cref{sec:train_obj}). We note the choice of star shape is not important --- any marker easily recognizable and unlikely to appear in natural images suffices (see~\cref{sec:star_ablation} for ablation of alternative click representations.) We then enforce all relevant information --- selection intent (object vs.~material), visual attributes (e.g., color, reflectance, roughness), and spatial relations --- to be encoded into a [SEG] token. An MLP projects this token from the textual embedding space into the visual feature space before passing it to the mask decoder, receiving no other conditioning regarding the intended selection. Finally, the decoder outputs a high-resolution (1024 $\times$ 1024) dense selection mask.

Our model is trained jointly on material and object data (\cref{sec:data}), allowing a rich, material-aware feature representation while retaining strong object selection capabilities. Importantly, we adapt the SAM-based mask decoder, trained on object-centric data, to also segment material-specific selections.
As a result, our model can select spatially disjoint regions that share the same material (e.g., all metal fixtures in a scene) while also supporting object selection.

During inference, the VLM encodes the user's selection intent based on the prompt, allowing the same image to yield different selections depending on the interaction, while the decoder simply consumes the [SEG] token and produces the mask. \cref{fig:main} illustrates this flexibility: for click-based material selection (using a standard prompt such as “segment everything made of the same material as where the star is”), the model selects both cups. For text-based selection with a prompt such as “the white ceramic in the front,” the model selects the specific cup referenced by the user.

\mysubsection{Training Objective}{train}\label{sec:train_obj}

Let $x_{\mathrm{img}}$ and $x_{\mathrm{txt}}$ denote the input image and the user text prompt, respectively. For click-based selection, we generate a visual prompt by overlaying a star marker at the click location, yielding $x^{*}_{\mathrm{img}}$. We intentionally represent clicks as a visual overlay rather than additional coordinate tokens, as this keeps the input modality uniform across tasks and allows the VLM to reason about the click in the same pixel space as other visual cues (e.g., nearby boundaries, texture, and context), which is important for material selection.

The input image ($x_{\mathrm{img}}$ or $x^{*}_{\mathrm{img}}$) is first encoded by a CLIP vision encoder \cite{radford2021learning}. The visual embedding is mapped into the LLM token embedding space via MLP, producing a fixed-length sequence $z_{\mathrm{img}}$. This projection serves two purposes: (i) it enables seamless fusion of visual and textual information within the LLM, and (ii) it avoids introducing a cross-modal module, keeping the architecture simple and compatible with pretrained LLM backbones.

The LLM processes the concatenated sequence $(z_{\mathrm{img}}, x_{\mathrm{txt}})$ and is trained to emit a newly introduced special token [SEG] whose hidden representation summarizes the full selection specification: the user's intent (object vs.~material), relevant visual attributes (e.g., color, reflectance, roughness), and spatial relations (e.g., ``in the front'', the click position). By bottlenecking all selection information into a single [SEG] embedding, we enforce that the VLM produces an explicit, task-relevant representation that is compatible with the interaction type (click vs.~text) and can be consumed by the decoder without additional hand-engineered prompts.

To generate the final mask, we feed the [SEG] embedding to SAM's prompt encoder \cite{kirillov2023sam}, and condition the SAM mask decoder to predict a dense, high-resolution selection mask. Importantly, the mask decoder receives no other task-specific conditioning beyond the [SEG] embedding; this design isolates language/intent understanding within the VLM, while leveraging SAM's strong mask prior for accurate boundary delineation.

We train the model with a multi-task objective that combines click-based selection, referring-text selection, and VQA:
\begin{equation*}
\mathcal{L}(x) = 
\lambda_1\mathcal{L}_{\mathrm{click}}(x^*_{\mathrm{img}}) + \lambda_2\mathcal{L}_{\mathrm{ref}}(x_{\mathrm{img}}, x_{\mathrm{txt}}) + \lambda_3\mathcal{L}_{\mathrm{vqa}}(x^*_{\mathrm{img}}, x_{\mathrm{txt}})
\end{equation*}
where $\mathcal{L}_{\mathrm{click}}$ and $\mathcal{L}_{\mathrm{ref}}$ denote click-based and text-based selection training, respectively, and $\mathcal{L}_{\mathrm{vqa}}$ denotes the VQA loss. We use the star-overlaid image $x^{*}_{\mathrm{img}}$ for click-based selection to explicitly convey the interaction point and align supervision with the user-provided click. We additionally use $x^{*}_{\mathrm{img}}$ for VQA to train the VLM to attend to the clicked region when answering questions, thereby aligning its visual reasoning with the same interaction cue used for click-based selection. In contrast, referring-text selection uses the original image $x_{\mathrm{img}}$, encouraging purely language-driven grounding without relying on auxiliary spatial markers.

The selection task losses ($\mathcal{L}_{\mathrm{click}}$ and $\mathcal{L}_{\mathrm{ref}}$) use the same combination of (i) token-level cross-entropy loss for language modeling (including the [SEG] token) and (ii) per-pixel mask supervision using binary cross-entropy (BCE) loss and DICE loss~\cite{dice}. We include both BCE and DICE to balance per-pixel optimization with robust region-level overlap under class imbalance. For VQA, we use token-level cross-entropy loss. Further details on the network architecture, training time, GPU requirements, model size, or inference latency are reported in Suppl.~\cref{sec:suppl_training_details}.

\begin{figure*}[t]
    \centering
    \includegraphics{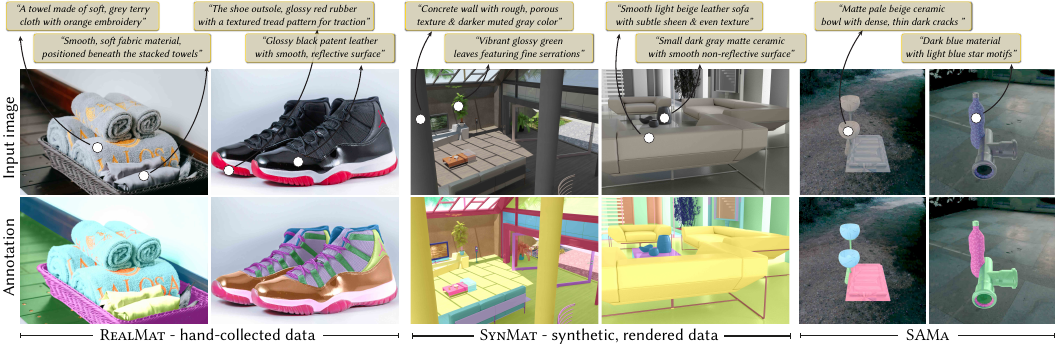}
    \caption{Example annotations. We show the images' dense, per-pixel material annotations overlaid in the second row. 
    We additionally show two versions of our generated text descriptions for these materials; one including the object (top row) and one solely focusing on the material (bottom row).}
    \label{fig:dataset}
\end{figure*}

\mysection{Dataset}{data}
Training a unified selection model requires material datasets with dense mask annotations and rich text descriptions. While existing material datasets provide high-quality mask annotations, they either lack textual descriptions~\cite{guerrero2025matselection, fischer2026sama, sharma2023materialistic}, are available only as flat material maps \cite{vecchio2024matsynth}, or are tied to a specific domain, e.g., fabrics \cite{deschaintre23_visual_fabric}. On the other hand, grounding and reasoning datasets (e.g., RefCOCOg \cite{mao2016generation}) have textual annotations but are not object centric and typically too short for fine-grained material selection. To address this, we collect new material data with dense mask annotations and develop a VLM-based pipeline to generate detailed text descriptions. 

Unlike recent reasoning segmentation datasets that use allusive queries (e.g., "select the food with the highest protein" when a steak is shown), our descriptions directly refer to visual material attributes with semantically aligned information --- we argue that users are unlikely to issue such indirect queries for selection tasks. Moreover, because our descriptions capture fine-grained visual details, the model learns associations between text and appearance, enabling generalization to diverse user prompts at inference time.

\mysubsection{Material Mask Data}{mat_mask}
We collect material mask data from both real and synthetic sources to capture natural diversity and precise, controlled annotations.

For real images, we collect $\sim$8K images from \href{https://www.pexels.com}{Pexels.com} and hand-annotate them, with the help of external users resulting in $\sim$49K material masks. We denote this dataset as \dataset{RealMat}.

For synthetic images, the subset of~\citet{guerrero2025matselection} is not suitable as materials are assigned irrespective of semantics --- e.g., a sofa with a checkerboard pattern of wood and stone which is unrealistic and incompatible for reasoning about real-world images. We instead render images with semantically correct materials --- e.g., a sofa made of leather or fabric --- using Blender and 132 scenes from \href{https://www.evermotion.org}{Evermotion.com} with pre-defined camera paths. We gather $\sim$5.5K images along with $\sim$55K material masks, and denote it \dataset{SynMat}.

Finally, we use SAMa~\cite{fischer2026sama}, consisting of $\sim$1.3K images and $\sim$3.3K material masks, which we denote as \dataset{SAMa}. Although its material assignments are not semantically meaningful, they are consistent across object parts. Both \dataset{SynMat} and \dataset{SAMa} consist of video frames across multiple viewpoints, providing varying mask shapes and material-light responses. 

This amounts to $\sim$104K material annotations from $\sim$15K images. We show in~\cref{tab:real_syn} that the real and synthetic datasets are complementary: training on both significantly improves performance on both evaluation sets compared to training on either subset. For further details on train and test splits, refer to Suppl.~\cref{sec:suppl_dataset_details}.

\mysubsection{Material Data Generation Pipeline}{data_gen}

Human annotation provides high-quality labels, but is prohibitively expensive at the scale required for our setting. To efficiently curate material descriptions, we use VLMs to generate candidate annotations and incorporate quality control through model-based verification and targeted human review. This hybrid pipeline enables scalable annotation while maintaining high-quality supervision.%

We describe the annotation and VQA generation processes below, followed by verification. Unless indicated otherwise, we use Qwen3-VL-235B-A22B-Thinking \cite{bai2025qwen3} as our annotation model.

\begin{figure}[h]
    \centering
    \includegraphics[width=\columnwidth]{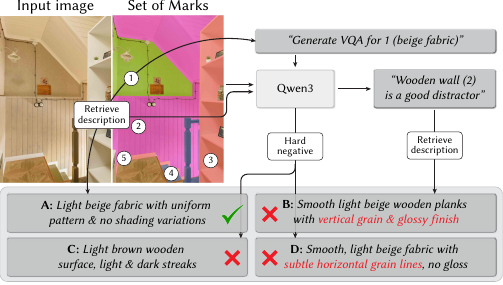}
    \caption{
        Example of VQA generation with hard negative mining. The VLM receives the SoM-overlaid image (denoted with numbered circles) in order to select the distractor and paraphrase the material description for both the answer and the negative candidate to create three distractors. The paraphrased parts are encoded in color (see \cref{sec:data_gen} for details).
    }
    \label{fig:vqa_gen}
\end{figure}

\paragraph{Description generation.}
We adopt Set-of-Marks (SoM) prompting \cite{yang2023set} to improve spatial grounding in vision–language reasoning. Given an input image, we provide the VLM with both a SoM-overlaid image (for an example, see \cref{fig:vqa_gen}) and a mask-overlaid image indicating which regions share the same material. 

For each marked region, we generate three types of descriptions. The first consists of a short material description augmented with an entity label, such as an object name or category (e.g., "chair"). The second combines a short material description with explicit spatial information --- either absolute image position (e.g., "bottom right corner") or relative to other objects (e.g., "above the table"). The third is a longer, self-contained material description that does not rely on contextual cues. We generate 6 variants of different lengths (10$\sim$50 words) which are randomly sampled during training.

\paragraph{Verification.} The generated descriptions occasionally suffer from incorrect grounding or instruction following --- e.g., including entity names in descriptions that should only contain material attributes. To address this, we perform verification using Qwen3-VL-235B-A22B-Thinking model as a verifier. In a manual audit of 500 sampled descriptions, a substantial amount of observed failures was fixed in the verification stage. Hence, we use the verified descriptions by default. However, the validation set is further inspected manually for accuracy in evaluating models. The filtering criteria are accuracy and unambiguity for both text-based segmentation and VQA. After manual filtering, we retain 1,797 out of 2,216 \dataset{RealMat} samples, 2,458 out of 3,072 \dataset{SynMat} samples, and 258 out of 352 \dataset{SAMa} samples, resulting in 4,513 out of 5,640 total ($\sim$80.0\%).

\paragraph{Visual Question Answering (VQA) Generation} Training on VQA encourages fine-grained material knowledge through reasoning in text. We formulate two variants of a four-way multiple-choice task.

We select a material in the image and retrieve its description from the previous stage as the answer. The distractors are constructed by sampling descriptions from other regions within the same image; if fewer than three distinct materials are present, we sample from other images. This requires the model to distinguish between visually present materials based on their descriptions.

The second variant introduces hard negative mining by generating visually plausible but incorrect alternatives --- e.g., changing "brown wood with dark streaks" to "horizontally grained light wood," as shown in \cref{fig:vqa_gen}. The negatives are generated for the answer and one sampled distractor. In \cref{sec:abldiscuss}, we show that training on VQA questions improves model performance and understanding.

\subsection{Training Data Composition}
Finally, to train a unified model for both object- and material-level selection, we incorporate publicly available object segmentation datasets. Notably, we show in~\cref{tab:mat_obj} that mixing in the object data does not deteriorate the model's performance on material selection.

We use the RefCOCO, RefCOCO+, and RefCOCOg~\cite{kazemzadeh-etal-2014-referitgame,mao2016generation} referring segmentation datasets for text-based object selection. 
For click-based object selection, we use EntitySeg~\cite{Qi_2023_EntitySeg} which consists of high-quality object selection masks from real-world images. 
This results in a total of $\sim$190K training samples with an approximate 1:1 ratio of material- and object-centric data, spanning diverse selection prompts and criteria. 
For more details on the datasets, see Suppl.~\cref{sec:suppl_dataset_details}.

\mysection{Evaluation}{evaluation}
In this section, we evaluate baselines along with MAOAM and report both quantitative and qualitative results across object and material selection tasks in both click- and text-based interactions.

\begin{table*}[t]
\centering
\caption{Comprehensive evaluation of \textbf{MAOAM} against state-of-the-art models. We report performance across material-specific datasets (\dataset{RealMat}, \dataset{SynMat}, \dataset{SAMa}) and object-centric datasets (RefCOCO, RefCOCO+, RefCOCOg and EntitySeg) using both text-based and click-based selection. Our model significantly outperforms existing methods on material selection datasets while retaining its object-selection capability. Materialistic does not allow text-based selection.}
\label{tab:main_results_combined}
\scriptsize
\setlength{\tabcolsep}{1.2pt} 
\begin{tabular*}{0.9\textwidth}{@{\extracolsep{\fill}} l | cccccc | cccccc | cccccc | cc @{}}
\toprule
& \multicolumn{6}{c|}{\textbf{Material (Text-based)}} & \multicolumn{6}{c|}{\textbf{Material (Click-based)}} & \multicolumn{6}{c|}{\textbf{Object (Text-based)}} & \multicolumn{2}{c}{\textbf{Object (Click-based)}} \\
\cmidrule(lr){2-7} \cmidrule(lr){8-13} \cmidrule(lr){14-19} \cmidrule(lr){20-21}
& \multicolumn{2}{c}{\dataset{RealMat}} & \multicolumn{2}{c}{\dataset{SynMat}} & \multicolumn{2}{c|}{\dataset{SAMa}} & \multicolumn{2}{c}{\dataset{RealMat}} & \multicolumn{2}{c}{\dataset{SynMat}} & \multicolumn{2}{c|}{\dataset{SAMa}} & \multicolumn{2}{c}{RefCOCO} & \multicolumn{2}{c}{RefCOCO+} & \multicolumn{2}{c|}{RefCOCOg} & \multicolumn{2}{c}{EntitySeg} \\
Method & mIoU $\uparrow$ & F1 $\uparrow$ & mIoU $\uparrow$ & F1 $\uparrow$ & mIoU $\uparrow$ & F1 $\uparrow$ & mIoU $\uparrow$ & F1 $\uparrow$ & mIoU $\uparrow$ & F1 $\uparrow$ & mIoU $\uparrow$ & F1 $\uparrow$ & mIoU $\uparrow$ & F1 $\uparrow$ & mIoU $\uparrow$ & F1 $\uparrow$ & mIoU $\uparrow$ & F1 $\uparrow$ & mIoU $\uparrow$ & F1 $\uparrow$ \\
\midrule
SAM3 & 0.263 & 0.293 & 0.224 & 0.253 & 0.068 & 0.074 & 0.538 & 0.624 & 0.505 & 0.599 & 0.623 & 0.710 & 0.433 & 0.472 & 0.329 & 0.364 & 0.422 & 0.456 & 0.664 & 0.748 \\
Materialistic & -- & -- & -- & -- & -- & -- & 0.524 & 0.709 & 0.680 & 0.884 & 0.535 & 0.718 & -- & -- & -- & -- & -- & -- & 0.147 & 0.256 \\
LISA & 0.332 & 0.396 & 0.319 & 0.383 & 0.215 & 0.259 & 0.129 & 0.163 & 0.094 & 0.124 & 0.056 & 0.074 & 0.732 & 0.797 & 0.638 & 0.702 & 0.665 & 0.733 & 0.209 & 0.249 \\
GLaMM & 0.349 & 0.415 & 0.328 & 0.396 & 0.260 & 0.305 & 0.185 & 0.238 & 0.159 & 0.210 & 0.101 & 0.129 & 0.616 & 0.692 & 0.521 & 0.597 & 0.603 & 0.679 & 0.364 & 0.423 \\
Sa2VA & 0.473 & 0.552 & 0.431 & 0.502 & 0.471 & 0.538 & 0.260 & 0.317 & 0.242 & 0.289 & 0.378 & 0.452 & 0.781 & 0.840 & 0.729 & 0.782 & 0.749 & 0.810 & 0.435 & 0.495 \\
MAOAM (Ours) & \best{0.740} & \best{0.798} & \best{0.608} & \best{0.669} & \best{0.685} & \best{0.754} & \best{0.808} & \best{0.868} & \best{0.766} & \best{0.835} & \best{0.747} & \best{0.823} & \best{0.809} & \best{0.895} & \best{0.744} & \best{0.853} & \best{0.778} & \best{0.875} & \best{0.821} & \best{0.901} \\
\bottomrule
\end{tabular*}
\end{table*}
\begin{figure*}[h]
    \centering
    \includegraphics[width=0.8\textwidth]{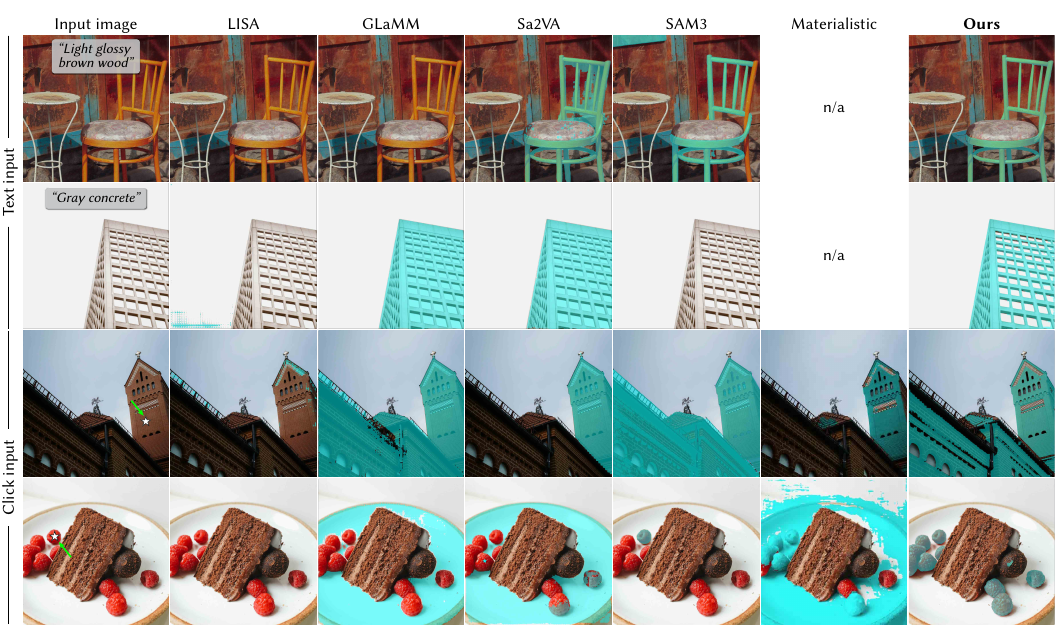}
    \caption{
        Material selection. We compare our method against baselines on a material selection task, both click- and text-based (first two and last two rows, respectively). LISA, GLaMM and SAM3 occasionally produce an empty mask when the selection criterion is too complicated or foreign to their vocabulary. Materialistic does not support text-based queries and is denoted n/a. 
    }
    \label{fig:mat_comparison}
    
    \centering
    \includegraphics[width=0.8\textwidth]{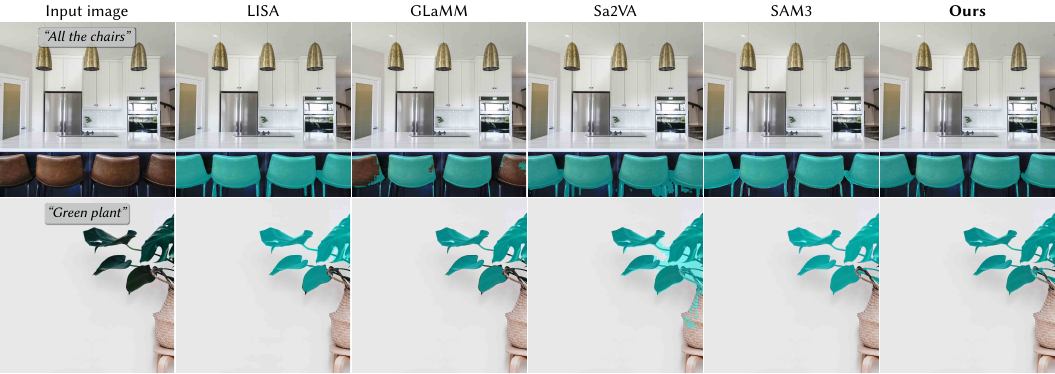}
    \caption{
        Object selection. We compare our method against several baselines on an object selection task. Materialistic neither supports text-based queries, nor object selection. Our method performs on par with the baselines, highlighting that reasoning about materials does not lead to deterioration on the object-level.
    }
    \label{fig:obj_comparison}
\end{figure*}

\mysubsection{Quantitative Evaluation}{eval_quantitative}

We evaluate MAOAM on material- and object-selection tasks using both text- and click-based inputs, where applicable. We compare against recent state-of-the-art VLM-based segmentation models, including GLaMM~\cite{rasheed2024glamm} (GranD-pretrained), Sa2VA~\cite{yuan2025sa2va}, LISA~\cite{lai2024lisa} (LISA-13B-llama2-v0-explanatory), SAM3~\cite{carion2025sam3segmentconcepts}, and the material-selection baseline Materialistic~\cite{sharma2023materialistic}. We report mean Intersection over Union (mIoU) and $\mathrm{F}_1$, the harmonic mean of precision and recall; higher is better for both.

\paragraph{Material Selection.}
The first two columns of \cref{tab:main_results_combined} report material selection performance on \dataset{RealMat}, \dataset{SynMat}, and \dataset{SAMa} across text- and click-based selection. 
MAOAM achieves substantial mIoU improvements over baselines (e.g., 67.5\% avg. mIoU over Sa2VA) in text-based selection, all of which do not perform well on material reasoning. For click-based selection, the performance gap is more pronounced as most baselines are not trained to process visual click prompts. MAOAM outperforms Materialistic which is trained for click-based material selection, by 35.5\% average mIoU. MAOAM performs strongly across all material datasets and interaction modalities, with high $\mathrm{F}_1$ scores indicating balanced selections.

\paragraph{Object Selection.}
The third and last block of \cref{tab:main_results_combined} reports performance on RefCOCO, RefCOCO+, RefCOCOg and EntitySeg for text- and click-based object selection. Albeit by a smaller margin (e.g., 33.9\%, 14.6\% and 3.2\% average mIoU improvement over GLaMM, LISA, and Sa2VA), MAOAM outperforms pretrained baselines, indicating joint training on materials does not degrade object selection.

\paragraph{Visual Question Answering.}
\cref{tab:eval_material_vqa} reports VQA performance on the two question variants described in \cref{sec:data_gen} (denoted Q1 and Q2).
We compare to Qwen2.5-VL-7B~\cite{Qwen2.5-VL}, a pretrained VLM with strong VQA performance. MAOAM achieves high accuracy on both, while Sa2VA and Qwen2.5-VL-7B, perform poorly, indicating that their ability to interpret visually differing stimuli from materials and their descriptions is limited.

Notably, MAOAM performs better on Q2 than Q1, opposite to the baselines, aligning with~\citet{cai2024temporalbench}’s finding that models with stronger domain understanding can recognize hard-negative variants as related. MAOAM's strong Q2 performance suggests it has acquired fine-grained material understanding, whereas the baselines are instead confused by the similar options. In~\cref{sec:ablation}, we show that incorporating VQA during training improves downstream selection performance despite the different objectives.

\mysubsection{Qualitative Evaluation}{qual_eval}
We further provide qualitative comparisons in~\cref{fig:mat_comparison,fig:obj_comparison}. MAOAM produces higher-precision masks with stronger grounding quality and generalizes to a wider range of material descriptions, while existing models often struggle when the selection criterion falls outside their object-centric vocabulary. We attribute this robustness to our description generation pipeline, which provides detailed material annotations with diverse reasoning patterns, including spatial relations. Although MAOAM is trained with longer, detailed material descriptions, the qualitative examples use shorter inference prompts, demonstrating generalization to more natural user queries. More examples are shown in Suppl.~\cref{fig:suppl_additional_examples}.

We next highlight several desirable properties of our model.

\begin{figure}[h]
    \centering
    \includegraphics[width=\linewidth]{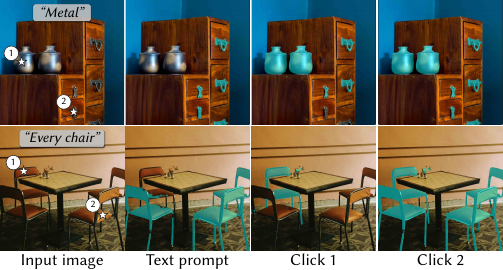}
    \caption{Emergent refinement. Starting from a text prompt, we show the selection result (material top, object bottom row) with an increasing number of interactions, i.e., more explicit guidance for the selection masks, closely resembling a realistic interaction scenario. Note that our model was never explicitly trained to handle both click- and text-prompts.}
    \label{fig:comparison1col}
    \vspace{-5mm}
\end{figure}

\begin{table}[t]
\centering
\caption{VQA performance on material datasets (\dataset{RealMat}, \dataset{SynMat}, \dataset{SAMa}). We report multiple-choice accuracy here.}
\label{tab:eval_material_vqa}
\small
\setlength{\tabcolsep}{5pt}
\begin{tabular*}{\columnwidth}{@{\extracolsep{\fill}} lllllll @{}}
\toprule
             & \multicolumn{2}{c}{\dataset{RealMat}} & \multicolumn{2}{c}{\dataset{SynMat}} & \multicolumn{2}{c}{\dataset{SAMa}} \\
\midrule
             & Q1 & Q2 & Q1 & Q2 & Q1 & Q2 \\
\midrule
Qwen2.5-VL-7B& 0.584 & 0.318 & 0.543 & 0.288 & 0.480 & 0.564 \\
Sa2VA        & 0.484 & 0.311 & 0.510 & 0.305  & 0.380 & 0.432 \\
MAOAM (Ours) & \best{0.858} & \best{0.974} & \best{0.795} & \best{0.979} & \best{0.749} & \best{0.858} \\
\bottomrule
\vspace{-6mm}
\end{tabular*}
\end{table}
\newpage
\paragraph{Emergent multimodal interaction.} Although MAOAM is trained with uni-modal prompts, combining text and click during inference improves mask quality (\cref{fig:comparison1col}). We highlight that this behavior emerges without explicit supervision, as our training data explicitly consists of uni-modal interactions.

\begin{figure}[h]
    \centering
    \includegraphics[width=0.8\linewidth]{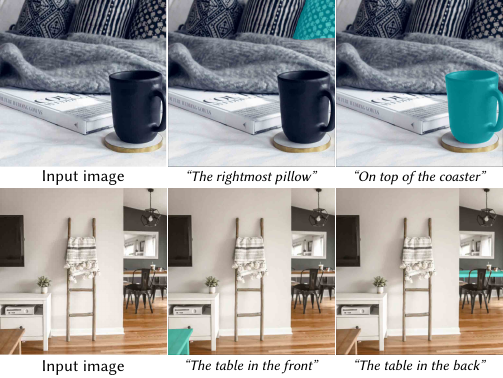}
    \caption{Spatial reasoning. Due to our training data (\cref{sec:data_gen}), MAOAM understands spatial queries relative to other entities and select accordingly.}
    \label{fig:spatialreasoning}
\end{figure}

\paragraph{Spatial and semantic reasoning.} MAOAM interprets spatial relations, leveraging the spatial descriptions in our training data (\cref{fig:spatialreasoning}). The model also handles prompts that span multiple objects --- e.g., "select everything made out of metal" --- and produces coherent masks across spatially disjoint regions.

\begin{figure}[h]
    \centering
    \begin{minipage}{\linewidth}
        \centering
        \includegraphics[width=\linewidth]{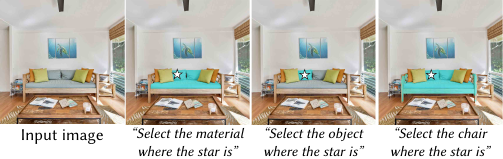}
        \captionof{figure}{Varying text prompts. The same visual prompt (star marker) is interpreted differently with varying text prompt.}
        \label{fig:varyingprompts}
    \end{minipage}

    \begin{minipage}{\linewidth}
        \centering
        \includegraphics[width=\linewidth]{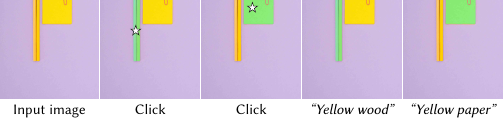}
        \captionof{figure}{Disambiguation. Although both objects are yellow, MAOAM infers the masks for both click and text prompts.}
        \label{fig:disambiguation}
    \end{minipage}
    \vspace{-3mm}
\end{figure}

\paragraph{Flexible Selection.} Given the same click location, different prompts yield object- or material-level selections (\cref{fig:varyingprompts}). For example, a click on a sofa can select either the entire sofa or the fabric depending on the task prompt. Our model also disambiguates between color and material when both could apply (\cref{fig:disambiguation}), demonstrating flexibility over various selection criteria and adherence to user instructions.

\begin{figure}[h]
    \centering
    \includegraphics[width=\linewidth]{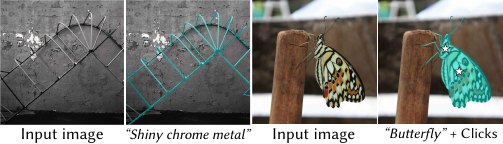}
    \caption{Mask quality. Our model performs well on intricate selection targets producing fine-grained masks. Images are from SAM-HQ \cite{ke2023samhq}.}
    \label{fig:finegrained}
    \vspace{-3mm}
\end{figure}

\paragraph{Mask quality.} MAOAM produces fine-grained, high-precision masks that capture detailed boundaries (\cref{fig:finegrained}). Unlike SAM-HQ~\cite{ke2023samhq}, our model achieves this without explicit guidance, showing that SAM decoder can produce fine-grained masks.

\begin{figure}[h]
    \centering
    \includegraphics[width=\linewidth]{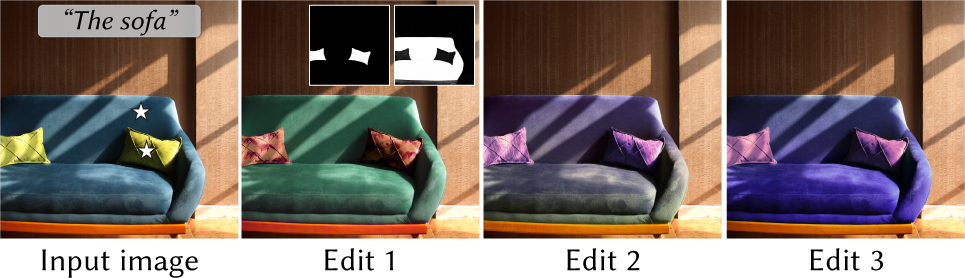}
    \caption{Editing. The selection output (masks displayed as insets) for click- and text-based queries can be used to edit materials in the image.}
    \label{fig:editing}
\end{figure}

\paragraph{Image Editing.} MAOAM produces high-quality, material-aware masks from both click- and text-based interactions, enabling real-world image editing workflows (\cref{fig:editing}) such as replacing all regions of one material with another material.

\mysection{Ablation \& Discussion}{abldiscuss}\label{sec:ablation}
All ablation experiments that require training are done on the GLaMM-based MAOAM model due to lower compute requirements.

\paragraph{Model choices in data generation} 
We ablate the effect of model size and verification in our description generation pipeline, comparing descriptions generated with Qwen3-VL-8B and Qwen3-VL-235B-A22B-Thinking in~\cref{tab:ablation_datagen}. Verification consistently improves metrics for descriptions generated with the 8B model across all three datasets. Although the gain is not as consistent for the descriptions generated with the 235B model, we empirically validated that the verification leads to more grounded descriptions and better instruction following, and hence set this configuration as default.

\begin{table}[h]
\centering
\caption{Description generation pipeline analysis. We report text-based material selection performance after training with the annotations produced by the respective pipeline configurations (rows). Columns Generate and Verify denote the model sizes used for description generation and verification.}
\label{tab:ablation_datagen}
\small
\setlength{\tabcolsep}{3pt}
\begin{tabular*}{\columnwidth}{@{\extracolsep{\fill}} ll cccccc @{}}
\toprule
& & \multicolumn{2}{c}{\dataset{RealMat}} & \multicolumn{2}{c}{\dataset{SynMat}} & \multicolumn{2}{c}{\dataset{SAMa}} \\
\cmidrule(lr){3-4} \cmidrule(lr){5-6} \cmidrule(lr){7-8}
Generate & Verify & mIoU $\uparrow$ & F1 $\uparrow$ & mIoU $\uparrow$ & F1 $\uparrow$ & mIoU $\uparrow$ & F1 $\uparrow$ \\
\midrule
8B   & --   & 0.644 & 0.712 & 0.532 & 0.604 & 0.626 & 0.703 \\
8B   & 235B & 0.657 & 0.726 & 0.554 & 0.628 & 0.651 & 0.721 \\
235B & --   & \best{0.676} & 0.744 & 0.575 & 0.649 & \best{0.661} & \best{0.738} \\
235B & 235B & 0.675 & \best{0.746} & \best{0.588} & \best{0.665} & 0.654 & 0.730 \\
\bottomrule
\end{tabular*}
\end{table}

\begin{figure}[h]
    \centering
    \includegraphics{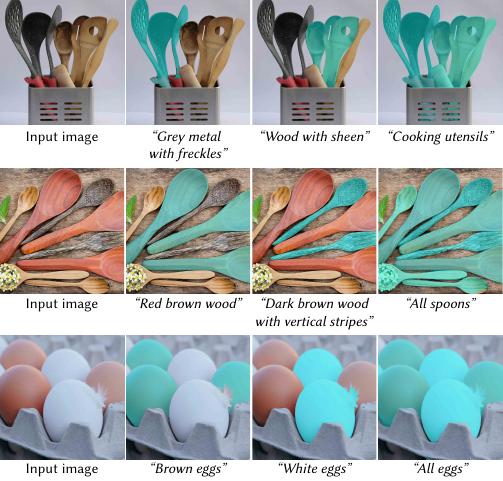}
    \caption{Joint object-material reasoning. MAOAM can switch between selecting material-specific subsets and full objects. In scenes containing the same object with different materials, material queries select the relevant subset, object queries select all instances, and joint material-object queries (e.g. \textit{``brown eggs''}) select instances satisfying both criteria.}  \label{fig:objects_and_materials}
\end{figure}

\paragraph{Training with objects helps} We ablate data composition by comparing MAOAM trained on material-only versus full mixed data (Materials + Objects). \cref{tab:mat_obj} reports material selection performance for both text- and click-based interactions. Despite joint object-material training, Materials + Objects remains competitive across all datasets and even improves some click-based results. This suggests that joint training preserves material understanding while adding object selection and joint object-material reasoning capabilities (\cref{fig:objects_and_materials}) at no cost to material performance.

\begin{table}[h]
\centering
\caption{Objects training data ablation. We report mIoU for text- and click-based material selection. Joint training on materials and objects maintains competitive performance compared to model trained on material only.}
\label{tab:mat_obj}
\small
\setlength{\tabcolsep}{4pt}
\begin{tabular*}{\columnwidth}{@{\extracolsep{\fill}} l cc cc cc @{}}
\toprule
& \multicolumn{2}{c}{\dataset{RealMat}} & \multicolumn{2}{c}{\dataset{SynMat}} & \multicolumn{2}{c}{\dataset{SAMa}} \\
\cmidrule(lr){2-3} \cmidrule(lr){4-5} \cmidrule(lr){6-7}
Training data & Text & Click & Text & Click & Text & Click \\
\midrule
Materials only & \best{0.675} & 0.756 & \best{0.588} & \best{0.730} & 0.654 & 0.730 \\
Materials + Objects  & 0.670 & \best{0.760} & 0.582 & 0.726 & \best{0.661} & \best{0.756} \\
\bottomrule
\end{tabular*}
\end{table}

\paragraph{Training with synthetic data helps.} A significant portion of our material training data (\dataset{RealMat}, \dataset{SynMat}, and \dataset{SAMa}) is synthetic. We ablate this by training on data subsets. \cref{tab:real_syn} shows that training on both datasets improves performance on both datasets by up to 9.15\%. Full training remains strongest, improving over single-dataset training by up to 21.49\%. This shows that synthetic data provide complementary supervision that transfers to real images, reducing reliance on human annotations which can be costly.

\begin{table}[h]
\centering
\caption{Material training data ablation. We report mIoU for text- and click-based material selection. Joint training on \dataset{RealMat} and \dataset{SynMat} improves performance on both evaluation sets. Training with all data performs best.}
\label{tab:real_syn}
\small
\setlength{\tabcolsep}{4pt}
\begin{tabular*}{\columnwidth}{@{\extracolsep{\fill}} l cc cc cc @{}}
\toprule
& \multicolumn{2}{c}{\dataset{RealMat}} & \multicolumn{2}{c}{\dataset{SynMat}} & \multicolumn{2}{c}{\dataset{SAMa}} \\
\cmidrule(lr){2-3} \cmidrule(lr){4-5} \cmidrule(lr){6-7}
Training data & Text & Click & Text & Click & Text & Click \\
\midrule
\dataset{RealMat} only                    & 0.579 & 0.694 & --    & --    & --    & --    \\
\dataset{SynMat} only                     & --    & --    & 0.484 & 0.683 & --    & --    \\
\dataset{SAMa} only                     & --    & --    & -- & -- & 0.595    & 0.665    \\
\dataset{RealMat} + \dataset{SynMat}                & 0.632 & 0.723 & 0.520 & 0.696 & --    & --    \\
All combined & \best{0.675} & \best{0.756} & \best{0.588} & \best{0.730} & \best{0.654} & \best{0.730} \\
\bottomrule
\end{tabular*}
\end{table}

\paragraph{Robustness to input text length} We generate six referring descriptions of varying length (10$\sim$50 words) and detail. The descriptions contain compositional reasoning and attribute combinations, enabling the diverse use cases shown in~\cref{fig:varyingprompts,fig:disambiguation,fig:spatialreasoning}. During inference, we observe that MAOAM generalizes to short, natural user prompts. To quantify MAOAM's robustness to input text length, we evaluate on all six different text prompts, group them by length (short, medium, long) and report the metrics in~\cref{tab:prompt_length}. Results show that performance remains stable across prompt lengths, with low variance on all three material benchmarks. This suggests that MAOAM does not rely on a fixed prompt template or length, but learns to ground the relevant material cues expressed in text.

\begin{table}[h]
\centering
\caption{Prompt length analysis. We report text-based material selection performance using short, medium, and long descriptions, with two prompt variants for each length. MAOAM remains robust across prompt lengths.}
\label{tab:prompt_length}
\small
\setlength{\tabcolsep}{3pt}
\begin{tabular*}{\columnwidth}{@{\extracolsep{\fill}} l cccccc @{}}
\toprule
& \multicolumn{2}{c}{\dataset{RealMat}} & \multicolumn{2}{c}{\dataset{SynMat}} & \multicolumn{2}{c}{\dataset{SAMa}} \\
\cmidrule(lr){2-3} \cmidrule(lr){4-5} \cmidrule(lr){6-7}
Prompt & mIoU $\uparrow$ & F1 $\uparrow$ & mIoU $\uparrow$ & F1 $\uparrow$ & mIoU $\uparrow$ & F1 $\uparrow$ \\
\midrule
Short  & 0.745 & 0.801 & 0.622 & 0.678 & 0.643 & 0.726 \\
Medium & 0.739 & 0.796 & 0.595 & 0.660 & 0.712 & 0.771 \\
Long   & 0.736 & 0.796 & 0.607 & 0.668 & 0.701 & 0.765 \\
\midrule
Mean   & 0.740 & 0.798 & 0.608 & 0.669 & 0.685 & 0.754 \\
Std    & 0.004 & 0.002 & 0.011 & 0.007 & 0.030 & 0.020 \\
\bottomrule
\end{tabular*}
\end{table}

\paragraph{Click representations.}\label{sec:star_ablation} We compare the star-overlay with alternative visual inputs. In~\cref{tab:click_representation}, ``Coordinates'' indicates providing the input coordinate to the mask decoder, whereas ``BBox'' indicates overlaying a bounding box on the target object. We note that ``BBox'' is how the original GLaMM model was trained. However, star-overlay achieves the best performance across all three material benchmarks. We hypothesize that the star-overlay provides the VLM with the grounding cue directly, allowing it to encode salient information in the [SEG] token. In contrast, ``Coordinates'' introduces the user query information at the decoding stage only, while ``BBox'' is less precise and may contain multiple different materials, confusing the model and leading to inferior performance. We emphasize the specific choice of a star is not essential; the marker only needs to be distinctive and easy for the model to recognize.

\begin{table}[h]
\centering
\caption{Click representation analysis. We report click-based material selection performance for different click representations. Star-overlay provides the best performance while preserving a unified input interface.}
\label{tab:click_representation}
\small
\setlength{\tabcolsep}{3pt}
\begin{tabular*}{\columnwidth}{@{\extracolsep{\fill}} l cccccc @{}}
\toprule
& \multicolumn{2}{c}{\dataset{RealMat}} & \multicolumn{2}{c}{\dataset{SynMat}} & \multicolumn{2}{c}{\dataset{SAMa}} \\
\cmidrule(lr){2-3} \cmidrule(lr){4-5} \cmidrule(lr){6-7}
Click representation & mIoU $\uparrow$ & F1 $\uparrow$ & mIoU $\uparrow$ & F1 $\uparrow$ & mIoU $\uparrow$ & F1 $\uparrow$ \\
\midrule
Star-overlay & \best{0.760} & \best{0.873} & \best{0.726} & \best{0.830} & \best{0.756} & \best{0.852} \\
Coordinates  & 0.755 & 0.843 & 0.714 & 0.812 & 0.727 & 0.822 \\
BBox         & 0.700 & 0.818 & 0.668 & 0.789 & 0.707 & 0.785 \\
\bottomrule
\end{tabular*}
\end{table}

\newpage

\begin{figure}[h]
    \centering
    \includegraphics{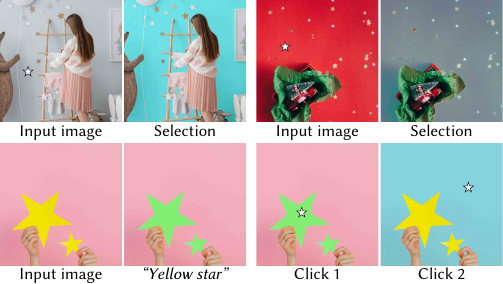}
    \caption{Star robustness. MAOAM remains robust when star-shaped objects appear in the image. In the second row, the text prompt is ``yellow star'' and no click is provided; MAOAM correctly segments the yellow stars, despite using colored star overlays only as small click markers.}
    \label{fig:supp_star_in_image}
\end{figure}

\paragraph{Presence of stars in the original image}
We overlay a $32\times32$ star at the click location, selecting its color from 10 candidates to maximize contrast with the region of interest. This naturally raises a concern when star-shaped objects appear in the image, as they could act as misleading grounding signals.~\cref{fig:supp_star_in_image} demonstrates MAOAM's robustness when star-shaped objects of varying sizes appear. In all cases, MAOAM correctly interprets the user-provided click and segments the intended region of interest. In the second row, we further test the text-only prompt \textit{``select the yellow star''} without any click input, and MAOAM correctly segments the yellow stars.

\paragraph{Joint material and object reasoning}{}
In~\cref{fig:objects_and_materials}, we provide additional examples showing why a single model capable of both material and object selection is useful. Each row contains objects from the same category with varying materials. The first two rows show that MAOAM can select a material-specific subset of the objects using material queries, while also selecting the full object set with an object query (e.g., cooking utensils). In the bottom row, the model further interprets joint material-object queries, such as \textit{``brown eggs''}, and adjusts its selection accordingly.

\begin{figure}[h]
    \centering
    \includegraphics[width=\linewidth]{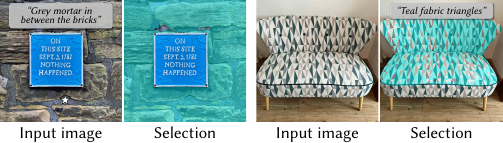}
    \caption{Limitations. For the first image, the model fails to distinguish the mortar from the bricks, while for the second image, the produced mask is inaccurate, presumably due to course resolution of the VLM image encoder.}
    \label{fig:limitations}
\end{figure}

\paragraph{Limitations} Our method can underperform in challenging cases. 
\cref{fig:limitations} highlights two failure modes: VLM reasoning and mask decoding. Reasoning is limited by the VLM backbone and may benefit from additional test-time compute via chain-of-thought tokens~\cite{kao2025think}. Mask quality is limited by the SAM decoder and could be improved with refinement modules~\cite{yao2024vitmatte}.

\mysection{Conclusion}{conclusion}
In this work, we present MAOAM, a unified framework for material- and object-selection with click- and text-based prompts. We propose a scalable, automatic annotation pipeline that enables us to generate a large corpus of rich material text descriptions for visual grounding. 
We demonstrate strong material selection performance while matching or outperforming object-centric segmentation methods.

\begin{acks}
This work was supported in part by NSF IIS2404180 and Institute of Information \& communications Technology Planning\& Evaluation (IITP) grant funded by the Korea government (MSIT) (No. 2022-0-00871, Development of AI Autonomy and Knowledge Enhancement for AI Agent Collaboration). The authors would like to thank Sudeep Katakol for his help in data generation and Zijun Wei, Yash Savani, and Soochahn Lee for helpful discussions.
\end{acks}

\bibliographystyle{ACM-Reference-Format}
\bibliography{main}

\clearpage
\appendix

\setcounter{section}{0}

\setcounter{figure}{0}

\setcounter{table}{0}

\renewcommand{\thesection}{S\arabic{section}}

\renewcommand{\thefigure}{S\arabic{figure}}

\renewcommand{\thetable}{S\arabic{table}}

\begin{center}
{\Large\bfseries Supplemental Material for MAOAM:\\[2pt]
Unified Object \& Material Selection with Vision-Language Models}
\end{center}
\vspace{1em}

\begin{figure*}[ht]
    \centering
    \includegraphics{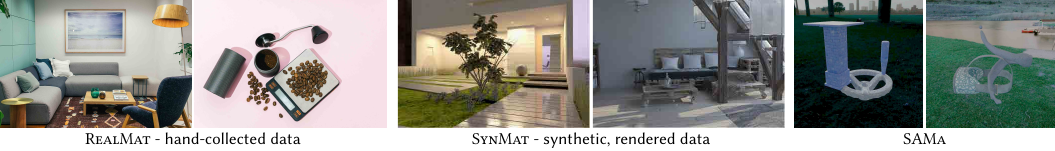}
    \caption{Additional examples from our material datasets.}
    \label{fig:supp_moredata}
\end{figure*}

In this supplementary material, we provide additional details on the datasets used to train our method as well as implementation and model details. We also provide more qualitative evaluation results and ablation studies that have been deferred due to limited space.

\section{Dataset Details}\label{sec:suppl_dataset_details}
We provide the number of source images and annotations for both train and validation splits for all datasets we use. 

\mysubsection{Material Datasets}{}

The material selection datasets provide click- and text-based prompts for material selection with precise material masks. 

\dataset{RealMat} consists of 7,848 images and 395 images in its training and validation sets, resulting in 46,646 and 2,214 material annotations for train and validation splits, respectively.

\dataset{SynMat} consists of 5,532 images and 352 images, which are frames sampled from videos, in its training and validation sets, which results in 54,315 and 3,071 material annotations for train and validation splits, respectively.

\dataset{SAMa} consists of 1,292 images and 141 images, which are also video frames, as its train and validation data. This results in 3,294 and 346 material annotations, for train and validation splits, respectively.

As a whole, our material dataset consists of $\sim$104K and $\sim$5.6K annotations for train and validation splits. \cref{fig:supp_moredata} provides more visual examples of our material dataset.

\mysubsection{Object and Entity Datasets}{}

\paragraph{RefCOCO}
We use the RefCOCO, RefCOCO+, and RefCOCOg datasets for text-based referring object selection. 
RefCOCO provides short, conversational referring expressions with relative spatial terms (e.g., "left of"). RefCOCO+ forbids location-based expressions, requiring appearance-based descriptions instead. RefCOCOg provides fewer but richer descriptions per object, with higher linguistic complexity. These datasets provide diverse object referring expressions that complement our material descriptions. RefCOCO contains 16,994 images, RefCOCO+ contains 16,992 images, and RefCOCOg contains 21,899 images. In total, the RefCOCO family provides approximately 56K training, 4.3K validation, and 5.6K test annotations. We use the official train, validation, and test splits.

\paragraph{EntitySeg}
The EntitySeg ~\cite{Qi_2023_EntitySeg} dataset provides referring prompts for click-based object selection.
It consists of $\sim$37K high-quality object selection masks from $\sim$8K real-world images collected from datasets such as COCO~\cite{coco}, ADE20K~\cite{ade20k} and LAION-400M~\cite{schuhmann2021laion400mopendatasetclipfiltered}. 

The original dataset contains small masks that are difficult to reliably annotate with star overlays. We filter out invalid masks and masks smaller than 0.3\% of the image area. After filtering, the dataset consists of 7,887 training images and 263 validation images, resulting in  $\sim$37K training and $\sim$2.7K validation annotations.

Combined, our material and object selection training data consists of $\sim$197K masks with varying selection criteria, material descriptions, and various orientations due to datasets that have been sampled from video frames.

\mysection{Training Details}{suppl_training_details}
We provide training details and hyperparameters for both backbone model configurations, as well as their architecture.

\mysubsection{Architecture and Hyperparameters}{hyperparam}
We train MAOAM on two backbone configurations: Sa2VA (Qwen2.5-VL-7B~\cite{Qwen2.5-VL} + SAM 2~\cite{ravi2024sam2}) and GLaMM \cite{rasheed2024glamm} (LLaVA-v1.5-7B~\cite{liu2024improved} + SAM~\cite{kirillov2023sam}). We list the hyperparameters below.

\paragraph{LLaVA-v1.5 and Qwen2.5-VL Architecture.} LLaVA-1.5~\cite{liu2024improved} pairs a frozen CLIP ViT-L/14 image encoder with a Vicuna LLM~\cite{vicuna2023} via a two-layer MLP projector that maps visual features into the LLM's token embedding space; the model is trained on image-text instruction data with the projector and LLM trainable. Qwen2.5-VL~\cite{Qwen2.5-VL} follows the same encoder–projector–LLM paradigm, with the addition of a native resolution ViT projected into the Qwen2.5 LLM backbone.

\paragraph{GLaMM Training and Inference}
We train from the GLaMM-GranD-Pretrained checkpoint for 15 epochs. We use a linear learning rate decay schedule with minimum learning rate $1e-6$ and warm-up for the first $100$ iterations. The initial learning rate is $2e-5$ for full VLM training and $3e-4$ for LoRA~\cite{hu2022lowrank} (rank 8, alpha 16). We use AdamW optimizer with $\beta_1=0.9, \beta_2 = 0.95$. For both models, we use the mask binary cross entropy loss and DICE loss for mask losses, and cross entropy loss for language modeling. We set $\lambda_{\mathrm{BCE}}=\lambda_{\mathrm{DICE}}=1.5$ and $\lambda_{\mathrm{CE}}=0.5$ for GLaMM training. 

For standard fine-tuning, we use a batch size of 4 and for LoRA fine-tuning, we use a batch size of 8. Since the VLM backbone is LLaVA, we train both the MLP adapter and the LLM, for both standard and LoRA fine-tuning cases. One epoch on our 190K Material and Object dataset takes approximately 8 hours on 8 A100 GPUs. During training, GLaMM-based MAOAM requires $\sim$50GB VRAM for training and $\sim$30GB VRAM during inference. Evaluating 1,000 images takes approximately one hour on 8 GPUs.

\paragraph{Sa2VA Training and Inference}
We train from the Sa2VA-7B model checkpoint trained with Qwen2.5-VL-7B as the VLM backbone and SAM 2 as the selection head. Qwen2.5-VL-7B requires significantly more GPU VRAM compared to LLaVA-v1.5, and hence we train the model with a batch size of 1, and gradient accumulation steps of 4. Similar to GLaMM, we use AdamW optimizer with $\beta_1 = 0.9, \beta_2 = 0.999$. We follow the default loss weights for Sa2VA, which are $\lambda_{\mathrm{BCE}}= 2.0, \lambda_{\mathrm{DICE}}=0.5$. We use LoRA training with LoRA rank of 128 and alpha 256, while keeping only the MLP adapter trainable, which is the default fine-tuning setup for Qwen2.5-VL. One epoch training of Sa2VA model on our 190K Material and Object data takes approximately 12 hours on eight A100 GPUs. During training, Sa2VA-based MAOAM requires $\sim$70GB VRAM for training and $\sim$50GB VRAM during inference. Evaluating 700 images takes approximately one hour on 8 GPUs.

For all of our experiments, we train GLaMM for 15 epochs and Sa2VA for 10 epochs, resulting in a comparable wall-clock time of approximately 120 hours on eight A100 GPUs. Finally, we note that MAOAM's inference is slightly faster than the baseline models, since it does not require additional modules to encode the positional information (e.g., GLaMM's region encoder), which we pass via the star-overlay in our framework.

\begin{table*}[t]
\centering
\caption{Comprehensive evaluation on object-centric datasets. We report performance across RefCOCO, RefCOCO+, and RefCOCOg on their respective validation and test splits (text-based object selection), as well as EntitySeg (click-based object selection). MAOAM consistently shows competitive performance despite being trained jointly on material dataset.}
\label{tab:object_centric_results_mega}
\scriptsize
\setlength{\tabcolsep}{2.5pt}
\begin{tabular*}{\textwidth}{@{\extracolsep{\fill}} l | cccccc | cccccc | cccc | cc @{}}
\toprule
& \multicolumn{6}{c|}{\textbf{RefCOCO (Text)}} & \multicolumn{6}{c|}{\textbf{RefCOCO+ (Text)}} & \multicolumn{4}{c|}{\textbf{RefCOCOg (Text)}} & \multicolumn{2}{c}{\textbf{EntitySeg (Click)}} \\
\cmidrule(lr){2-7} \cmidrule(lr){8-13} \cmidrule(lr){14-17} \cmidrule(lr){18-19}
& \multicolumn{2}{c}{Val} & \multicolumn{2}{c}{TestA} & \multicolumn{2}{c|}{TestB} & \multicolumn{2}{c}{Val} & \multicolumn{2}{c}{TestA} & \multicolumn{2}{c|}{TestB} & \multicolumn{2}{c}{Val} & \multicolumn{2}{c|}{Test} & \multicolumn{2}{c}{Val} \\
Method & mIoU $\uparrow$ & F1 $\uparrow$ & mIoU $\uparrow$ & F1 $\uparrow$ & mIoU $\uparrow$ & F1 $\uparrow$ & mIoU $\uparrow$ & F1 $\uparrow$ & mIoU $\uparrow$ & F1 $\uparrow$ & mIoU $\uparrow$ & F1 $\uparrow$ & mIoU $\uparrow$ & F1 $\uparrow$ & mIoU $\uparrow$ & F1 $\uparrow$ & mIoU $\uparrow$ & F1 $\uparrow$ \\
\midrule
SAM3 & 0.433 & 0.472 & 0.374 & 0.411 & 0.459 & 0.493 & 0.329 & 0.364 & 0.350 & 0.387 & 0.325 & 0.355 & 0.422 & 0.456 & 0.387 & 0.418 & 0.664 & 0.748 \\
LISA & 0.732 & 0.797 & 0.761 & 0.822 & 0.696 & 0.767 & 0.638 & 0.702 & 0.720 & 0.779 & 0.589 & 0.662 & 0.665 & 0.733 & 0.689 & 0.759 & 0.209 & 0.249 \\
GLaMM & 0.616 & 0.692 & 0.655 & 0.736 & 0.620 & 0.692 & 0.521 & 0.597 & 0.525 & 0.604 & 0.480 & 0.551 & 0.603 & 0.679 & 0.620 & 0.698 & 0.364 & 0.423 \\
Sa2VA & 0.781 & 0.840 & 0.819 & 0.874 & 0.765 & 0.826 & 0.729 & 0.782 & 0.787 & 0.842 & 0.671 & 0.731 & 0.749 & 0.810 & 0.758 & 0.819 & 0.435 & 0.495 \\
MAOAM (Ours) & \best{0.809} & \best{0.895} & \best{0.835} & \best{0.910} & \best{0.770} & \best{0.870} & \best{0.744} & \best{0.853} & \best{0.823} & \best{0.903} & \best{0.715} & \best{0.834} & \best{0.778} & \best{0.875} & \best{0.796} & \best{0.886} & \best{0.821} & \best{0.901} \\
\bottomrule
\end{tabular*}
\end{table*}

\mysubsection{Detailed Task Formulation}{task_form}

\paragraph{Multi-task Training} Each data point in our material selection data consists of three tasks: click- and text-based selection, and VQA questions. Hence, we formulate the loss function as a multi-task loss, where the click-selection, text-selection, and VQA tasks are weighted $\lambda_{\mathrm{click}} = 0.4, \lambda_{\mathrm{ref}} = 0.4, \lambda_{\mathrm{vqa}} = 0.2$. For single task case, i.e., RefCOCO or EntitySeg, where only $\mathcal{L}_{\mathrm{ref}}$ and $\mathcal{L}_{\mathrm{click}}$ are computed, respectively, we set the loss weights to $1$.

\paragraph{Star overlay}\label{apdx:star_overlay} During training, we randomly place 1-5 stars of size $32\times32$ pixels on the input image ($1024\times1024$). We empirically observe that the model is sensitive to star locations, especially for thin selection areas. To ensure the star overlay provides a clear signal while also making the model robust to boundary cases, we erode the target area's binary mask using MaxPool2D with kernel size $r=8$ to ensure most stars are included in the target area. When sampling multiple stars, we define boundary regions via erosion and place stars on boundary pixels with probability 0.5, and ensure that the stars are sufficiently far away from each other. 

Finally, the color of the star overlay is dynamically determined to have the highest contrast from the region the star is being overlaid on to (for visualizations in the paper, we use a default white star for visibility). In case we place multiple stars, the first star's color is used throughout. There are a total of 10 star colors, and we use the same logic during model inference as well.

\paragraph{System prompt for material selection}
We provide example prompts used for each task during training. All material-related prompts include a task prompt that clarifies the distinction between material and appearance variations. For each template, we generate about 3 to 7 paraphrased variants for diversity during training.\newpage

\paragraph{Click-based material selection.}
\begin{quote}
\texttt{Can you segment all pixels with the same material where the <COLOR> star is located? [MATERIAL\_PROMPT]}
\end{quote}

\paragraph{Text-based material selection.}
\begin{quote}
\texttt{Segment every region that has the material described below. [MATERIAL\_PROMPT]}\\
\texttt{Description: <DESCRIPTION>}
\end{quote}

\paragraph{VQA}
\begin{quote}
\texttt{Which of the following options best describes the material where the <COLOR> star is located? [MATERIAL\_PROMPT]}
\end{quote}

\paragraph{Answer templates.}
\begin{quote}
\texttt{Sure, the segmentation result is [SEG].}
\end{quote}

\paragraph{Material prompt.}
\begin{quote}
\texttt{Regions with same base material but different colors are considered as different materials. However, regions with different lighting, shading or shadows are considered as the same material.}
\end{quote}

\noindent <COLOR> is replaced with the star overlay color (e.g., red, cyan), and <DESCRIPTION> is replaced with the material description.

\paragraph{System prompt for object selection.}
For RefCOCO datasets, we follow the original implementation. The dataset contains short object descriptions, and the full question is formatted as:
\begin{quote}
\texttt{What is the <DESCRIPTION> in this image? Please output segmentation mask.}
\end{quote}
\noindent where \texttt{<DESCRIPTION>} is replaced with the referring expression (e.g., ``left side monitor'').

For EntitySeg, each datapoint contains a class name for the corresponding mask. To formulate instance segmentation, we provide the spatial location via a star overlay:
\begin{quote}
\texttt{Can you segment the <CLASS\_NAME> that contains the <COLOR> star?}
\end{quote}
\noindent where \texttt{<CLASS\_NAME>} is replaced with the object (e.g., chair, person).

In our experiments, we find that fine-tuning the entire VLM achieves significantly higher performance than using a low-rank adapter (LoRA). This is likely because our training data includes rich, fine-grained material descriptions, requiring the model to significantly adapt its visual-language representations.

\begin{figure*}[h]
    \centering \includegraphics[width=0.85\linewidth]{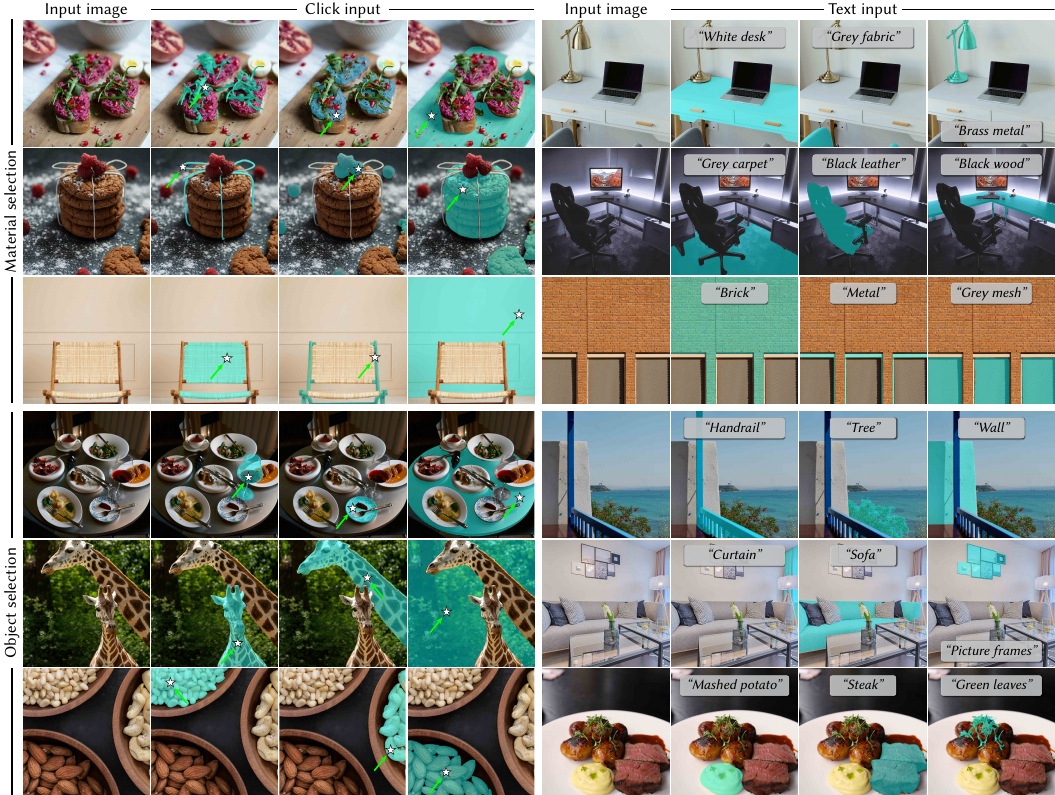}
    \caption{Additional examples of our method on material selection (first three rows) and object selection (last three rows). We show both click-based queries (first four columns) and prompt-based queries (last four columns).}
    \label{fig:suppl_additional_examples}
\end{figure*}
\begin{table*}[t]
\centering
\caption{Comparison of LoRA and full VLM fine-tuning for material understanding. We evaluate the performance across text-based selection, click-based selection, and VQA. Full VLM fine-tuning consistently outperforms the LoRA adaptation across all text-based tasks, including selection and VQA, while LoRA adaptation demonstrates comparable, and in certain cases better metrics on click-based selection.}
\label{tab:cmp_lora_vlm_mega}
\scriptsize
\setlength{\tabcolsep}{2.2pt}
\begin{tabular*}{\textwidth}{@{\extracolsep{\fill}} l | cccccc | cccccc | cccccc @{}}
\toprule
& \multicolumn{6}{c|}{\textbf{Material (Text-based Selection)}} & \multicolumn{6}{c|}{\textbf{Material (Click-based Selection)}} & \multicolumn{6}{c}{\textbf{Material (VQA)}} \\
\cmidrule(lr){2-7} \cmidrule(lr){8-13} \cmidrule(lr){14-19}
& \multicolumn{2}{c}{\dataset{RealMat}} & \multicolumn{2}{c}{\dataset{SynMat}} & \multicolumn{2}{c|}{\dataset{SAMa}} & \multicolumn{2}{c}{\dataset{RealMat}} & \multicolumn{2}{c}{\dataset{SynMat}} & \multicolumn{2}{c|}{\dataset{SAMa}} & \multicolumn{2}{c}{\dataset{RealMat} (Acc $\uparrow$)} & \multicolumn{2}{c}{\dataset{SynMat} (Acc $\uparrow$)} & \multicolumn{2}{c}{\dataset{SAMa} (Acc $\uparrow$)} \\
Method & mIoU $\uparrow$ & F1 $\uparrow$ & mIoU $\uparrow$ & F1 $\uparrow$ & mIoU $\uparrow$ & F1 $\uparrow$ & mIoU $\uparrow$ & F1 $\uparrow$ & mIoU $\uparrow$ & F1 $\uparrow$ & mIoU $\uparrow$ & F1 $\uparrow$ & Q1 & Q2 & Q1 & Q2 & Q1 & Q2 \\
\midrule
LoRA & 0.614 & 0.690 & 0.550 & 0.627 & 0.601 & 0.682 & 0.694 & 0.774 & 0.682 & 0.769 & 0.711 & 0.794 & 0.732 & 0.649 & 0.688 & 0.635 & 0.552 & 0.364 \\
VLM & \best{0.670} & \best{0.739} & \best{0.582} & \best{0.658} & \best{0.661} & \best{0.739} & \best{0.760} & \best{0.830} & \best{0.726} & \best{0.806} & \best{0.756} & \best{0.832} & \best{0.818} & \best{0.976} & \best{0.775} & \best{0.983} & \best{0.693} & \best{0.847} \\
\bottomrule
\end{tabular*}
\end{table*}

\begin{table*}[t]
\centering
\caption{Ablation study on multi-task training. We compare three configurations: \textit{Click}, \textit{Click+Text}, and \textit{All} (our multi-task training). We report grounding performance (mIoU and F1) and VQA Accuracy across all material datasets. We verify that introducing VQA questions helps improve the metrics on text-based selection task. The mixed results in click-based selection also signal that the three different task formulations complement each other. Note that VQA performance on \textit{Click} and \textit{Click+Text} models could not be measured because the models are not trained for VQA tasks.  All models are trained on material data only, for 10 epochs.}
\label{tab:ablation_training_data_mega}
\scriptsize
\setlength{\tabcolsep}{2.2pt}
\begin{tabular*}{\textwidth}{@{\extracolsep{\fill}} l | cccccc | cccccc | cccccc @{}}
\toprule
& \multicolumn{6}{c|}{\textbf{Material (Text-based Selection)}} & \multicolumn{6}{c|}{\textbf{Material (Click-based Selection)}} & \multicolumn{6}{c}{\textbf{Material (VQA)}} \\
\cmidrule(lr){2-7} \cmidrule(lr){8-13} \cmidrule(lr){14-19}
& \multicolumn{2}{c}{\dataset{RealMat}} & \multicolumn{2}{c}{\dataset{SynMat}} & \multicolumn{2}{c|}{\dataset{SAMa}} & \multicolumn{2}{c}{\dataset{RealMat}} & \multicolumn{2}{c}{\dataset{SynMat}} & \multicolumn{2}{c|}{\dataset{SAMa}} & \multicolumn{2}{c}{\dataset{RealMat} (Acc $\uparrow$)} & \multicolumn{2}{c}{\dataset{SynMat} (Acc $\uparrow$)} & \multicolumn{2}{c}{\dataset{SAMa} (Acc $\uparrow$)} \\
Strategy & mIoU $\uparrow$ & F1 $\uparrow$ & mIoU $\uparrow$ & F1 $\uparrow$ & mIoU $\uparrow$ & F1 $\uparrow$ & mIoU $\uparrow$ & F1 $\uparrow$ & mIoU $\uparrow$ & F1 $\uparrow$ & mIoU $\uparrow$ & F1 $\uparrow$ & Q1 & Q2 & Q1 & Q2 & Q1 & Q2 \\
\midrule
Click     & 0.093 & 0.120 & 0.067 & 0.089 & 0.265 & 0.323 & 0.757 & 0.829 & \best{0.735} & \best{0.814} & 0.745 & 0.822 & -- & -- & -- & -- & -- & -- \\
Click + Text  & 0.670 & 0.738 & \best{0.589} & \best{0.666} & \best{0.655} & 0.730 & 0.754 & 0.825 & 0.723 & 0.803 & 0.729 & 0.807 & -- & -- & -- & -- & -- & -- \\
All         & \best{0.675} & \best{0.746} & 0.588 & 0.665 & 0.654 & \best{0.730} & \best{0.756} & \best{0.828} & 0.730 & 0.810 & \best{0.730} & \best{0.809} & \best{0.833} & \best{0.970} & \best{0.771} & \best{0.979} & \best{0.705} & \best{0.835} \\
\bottomrule
\end{tabular*}
\end{table*}

\mysection{Full Evaluation Results}{Eval}
In this section, we list all validation results that have been deferred due to limited space. Throughout, MAOAM refers to our model trained from Sa2VA model for 10 epochs on material and object data. We report results on material selection and object selection with two distinct prompting modalities: text- and click-based, and Visual Question Answering (VQA), when applicable.

\paragraph{Material-Centric Understanding}
Tables~2 and~3 in the main text provide a comprehensive evaluation on material datasets: \dataset{RealMat}, \dataset{SynMat}, and \dataset{SAMa}. We evaluate material segmentation from click- and text-prompts, as well as reasoning via two VQA question types (Q1: sampling-based; Q2: hard-negative mining). MAOAM substantially outperforms existing models such as GLaMM and Sa2VA, which struggle with fine-grained material properties.

\paragraph{Object-Centric Grounding}
To ensure that our material-specific tuning does not degrade general-purpose capabilities, we provide full results on standard benchmarks in Table~\ref{tab:object_centric_results_mega}. This includes the validation and test splits for \dataset{RefCOCO}, \dataset{RefCOCO+}, and \dataset{RefCOCOg}, alongside click-based selection on \dataset{EntitySeg}. The results indicate that MAOAM not only preserves but often improves upon the performance of the base Sa2VA model in traditional referring expression segmentation tasks.

\begin{table*}[t]
\centering
\caption{Comparison between GLaMM and Sa2VA models' performance on material datasets, after being trained on our material and object dataset. Sa2VA (MAOAM) outperforms GLaMM by a large margin on both text- and click-based selection, despite being trained for fewer epochs. The two models show comparable performance on VQA.}
\label{tab:cmp_maoam_glamm_material}
\scriptsize
\setlength{\tabcolsep}{2.2pt}
\begin{tabular*}{\textwidth}{@{\extracolsep{\fill}} l | cccccc | cccccc | cccccc @{}}
\toprule
& \multicolumn{6}{c|}{\textbf{Material (Text-based Selection)}} & \multicolumn{6}{c|}{\textbf{Material (Click-based Selection)}} & \multicolumn{6}{c}{\textbf{Material (VQA)}} \\
\cmidrule(lr){2-7} \cmidrule(lr){8-13} \cmidrule(lr){14-19}
& \multicolumn{2}{c}{\dataset{RealMat}} & \multicolumn{2}{c}{\dataset{SynMat}} & \multicolumn{2}{c|}{\dataset{SAMa}} & \multicolumn{2}{c}{\dataset{RealMat}} & \multicolumn{2}{c}{\dataset{SynMat}} & \multicolumn{2}{c|}{\dataset{SAMa}} & \multicolumn{2}{c}{\dataset{RealMat} (Acc $\uparrow$)} & \multicolumn{2}{c}{\dataset{SynMat} (Acc $\uparrow$)} & \multicolumn{2}{c}{\dataset{SAMa} (Acc $\uparrow$)} \\
Method & mIoU $\uparrow$ & F1 $\uparrow$ & mIoU $\uparrow$ & F1 $\uparrow$ & mIoU $\uparrow$ & F1 $\uparrow$ & mIoU $\uparrow$ & F1 $\uparrow$ & mIoU $\uparrow$ & F1 $\uparrow$ & mIoU $\uparrow$ & F1 $\uparrow$ & Q1 & Q2 & Q1 & Q2 & Q1 & Q2 \\
\midrule
GLaMM & 0.670 & 0.739 & 0.582 & 0.658 & 0.661 & 0.739 & 0.760 & 0.830 & 0.726 & 0.806 & \best{0.756} & \best{0.832} & 0.818 & \best{0.976} & 0.775 & \best{0.983} & 0.693 & 0.847 \\
Sa2VA (MAOAM) & \best{0.740} & \best{0.798} & \best{0.608} & \best{0.669} & \best{0.685} & \best{0.754} & \best{0.808} & \best{0.868} & \best{0.766} & \best{0.835} & 0.747 & 0.823 & \best{0.858} & 0.974 & \best{0.795} & 0.979 & \best{0.749} & \best{0.858} \\
\bottomrule
\end{tabular*}
\end{table*}
\begin{table*}[t]
\centering
\caption{Comparison between GLaMM and Sa2VA models' performance on object datasets, after being trained on our material and object dataset. Sa2VA (MAOAM) outperforms GLaMM by a large margin across all text- and click-based object selection, despite being trained for less amount of epochs.}
\label{tab:cmp_maoam_glamm_object}
\scriptsize
\setlength{\tabcolsep}{2.5pt}
\begin{tabular*}{\textwidth}{@{\extracolsep{\fill}} l | cccccc | cccccc | cccc | cc @{}}
\toprule
& \multicolumn{6}{c|}{\textbf{\dataset{RefCOCO} (Text)}} & \multicolumn{6}{c|}{\textbf{\dataset{RefCOCO+} (Text)}} & \multicolumn{4}{c|}{\textbf{\dataset{RefCOCOg} (Text)}} & \multicolumn{2}{c}{\textbf{\dataset{EntitySeg} (Click)}} \\
\cmidrule(lr){2-7} \cmidrule(lr){8-13} \cmidrule(lr){14-17} \cmidrule(lr){18-19}
& \multicolumn{2}{c}{Val} & \multicolumn{2}{c}{TestA} & \multicolumn{2}{c|}{TestB} & \multicolumn{2}{c}{Val} & \multicolumn{2}{c}{TestA} & \multicolumn{2}{c|}{TestB} & \multicolumn{2}{c}{Val} & \multicolumn{2}{c|}{Test} & \multicolumn{2}{c}{Val} \\
Method & mIoU $\uparrow$ & F1 $\uparrow$ & mIoU $\uparrow$ & F1 $\uparrow$ & mIoU $\uparrow$ & F1 $\uparrow$ & mIoU $\uparrow$ & F1 $\uparrow$ & mIoU $\uparrow$ & F1 $\uparrow$ & mIoU $\uparrow$ & F1 $\uparrow$ & mIoU $\uparrow$ & F1 $\uparrow$ & mIoU $\uparrow$ & F1 $\uparrow$ & mIoU $\uparrow$ & F1 $\uparrow$ \\
\midrule
GLaMM & 0.772 & 0.834 & 0.798 & 0.857 & 0.754 & 0.819 & 0.719 & 0.778 & 0.761 & 0.819 & 0.677 & 0.740 & 0.727 & 0.792 & 0.734 & 0.799 & 0.777 & 0.846 \\
Sa2VA (MAOAM) & \best{0.809} & \best{0.895} & \best{0.835} & \best{0.910} & \best{0.770} & \best{0.870} & \best{0.744} & \best{0.853} & \best{0.823} & \best{0.903} & \best{0.715} & \best{0.834} & \best{0.778} & \best{0.875} & \best{0.796} & \best{0.886} & \best{0.821} & \best{0.901} \\
\bottomrule
\end{tabular*}
\end{table*}

\mysection{Discussion and Ablation Studies}{ablation}
In this section, we perform further discussion and ablation studies. All models used in ablation studies are initialized from GLaMM checkpoints and trained for 10 epochs, unless mentioned otherwise.

\paragraph{LoRA vs. standard fine-tuning}{}
While the default configuration of GLaMM utilizes LoRA with rank 8 and alpha 16, our experiments indicate that LoRA yields significantly lower performance compared to standard fine-tuning on text-based selection tasks, as shown in \cref{tab:cmp_lora_vlm_mega}. Interestingly, the LoRA-trained model demonstrates comparable or superior metrics in click-based selection. This suggests that low-rank adaptation is sufficient for processing local spatial information to produce accurate masks. However, standard fine-tuning of the LLM is clearly advantageous to interpret intricate and long material descriptions and align them with visual features. This performance gap is the most evident in VQA, where the model must reason within the text space to distinguish correct material attributes. For this reason, we follow standard fine-tuning as the default training strategy for GLaMM. For Sa2VA training, we follow the default setting (LoRA rank of 128) due to VRAM requirements.

\paragraph{Effect of multi-task training}{}
We evaluate the impact of our multi-task objective on material understanding by ablating three training configurations: (i) Click-only, which trains the model only on click-based selection; (ii) Click+Text, which combines click- and text-based selection; and (iii) All, our full multi-task framework with click-based selection, text-based selection, and VQA. All three models have been trained exclusively on material datasets.

As shown in \cref{tab:ablation_training_data_mega}, the results demonstrate a clear synergistic effect across tasks. The Click-only baseline performs well on spatial localization but cannot generalize to text-based queries. Adding text-based selection (Click+Text) restores referring performance, but the full multi-task configuration provides the largest gains. Specifically, including VQA not only complements text-based selection but also improves click-based selection, achieving the highest mIoU on RealMat over the three models. The varied results on click-based selection suggest that the three task formulations provide complementary supervision, resulting in a more robust model.

\paragraph{GLaMM vs Sa2VA}{}

We compare two backbone configurations: GLaMM (LLaVA-v1.5 + SAM) and Sa2VA (Qwen2.5-VL-7B + SAM-2) after training on our material and object data. Specifically, we train GLaMM for 15 epochs and Sa2VA for 10 epochs, resulting in comparable wall-clock time.

\cref{tab:cmp_maoam_glamm_material} and \cref{tab:cmp_maoam_glamm_object} report performance on material and object datasets, respectively. Sa2VA (MAOAM) substantially outperforms GLaMM across text- and click-based interactions, on both material and object selection, despite being trained for fewer epochs. VQA performance is comparable between the two models, with GLaMM slightly outperforming on some splits and MAOAM on others.

These results suggest that the more recent VLM backbone (Qwen2.5-VL) can better align complex text queries with visual-semantic representations that benefit both material and object selection. We therefore use Sa2VA as our primary model (MAOAM) but mainly use GLaMM for ablation studies due to lower computational cost.

We note that the Sa2VA-based variant yields higher quantitative metrics, while the GLaMM-based variant generalizes better and is more robust during inference. Hence, quantitative results are from the former and qualitative examples from the latter.

\paragraph{Data scaling.} We further evaluate the scalability of our data generation framework by varying both the amount of training data and the number of training epochs. As shown in~\cref{tab:data_scale}, training with only 50\% of randomly sampled material data remains competitive with full-scale training across all three material benchmarks. In~\cref{tab:epoch_analysis}, we also report performance after 5, 10, and 15 epochs of material training. While performance improves with longer training, 5--10 epochs of training already achieve competitive results. Together Table~4 in the main paper, these results suggest practical flexibility in the data generation and training pipeline, depending on the available compute budget.

\begin{table}[h]
\centering
\caption{Data scale analysis. We report mIoU for text- and click-based material selection when training with the full material dataset and a 50\% randomly subsampled version. Half-scale training remains competitive with full-scale training across all three material benchmarks.}
\label{tab:data_scale}
\small
\setlength{\tabcolsep}{4pt}
\begin{tabular*}{\columnwidth}{@{\extracolsep{\fill}} l cc cc cc @{}}
\toprule
& \multicolumn{2}{c}{\dataset{RealMat}} & \multicolumn{2}{c}{\dataset{SynMat}} & \multicolumn{2}{c}{\dataset{SAMa}} \\
\cmidrule(lr){2-3} \cmidrule(lr){4-5} \cmidrule(lr){6-7}
Training data & Text & Click & Text & Click & Text & Click \\
\midrule
Half & 0.641 & 0.732 & 0.570 & 0.726 & 0.639 & 0.727 \\
Full & \best{0.675} & \best{0.756} & \best{0.588} & \best{0.730} & \best{0.654} & \best{0.730} \\
\bottomrule
\end{tabular*}
\end{table}

\begin{table}[h]
\centering
\caption{Training epoch analysis. We report mIoU for text- and click-based material selection after 5, 10, and 15 epochs of training. Performance improves with longer training, while 5--10 epochs provide competitive results.}
\label{tab:epoch_analysis}
\small
\setlength{\tabcolsep}{4pt}
\begin{tabular*}{\columnwidth}{@{\extracolsep{\fill}} l cc cc cc @{}}
\toprule
& \multicolumn{2}{c}{\dataset{RealMat}} & \multicolumn{2}{c}{\dataset{SynMat}} & \multicolumn{2}{c}{\dataset{SAMa}} \\
\cmidrule(lr){2-3} \cmidrule(lr){4-5} \cmidrule(lr){6-7}
Training & Text & Click & Text & Click & Text & Click \\
\midrule
5 epochs  & 0.640 & 0.717 & 0.565 & 0.702 & 0.625 & 0.675 \\
10 epochs & 0.656 & 0.741 & 0.586 & 0.725 & \best{0.662} & 0.713 \\
15 epochs & \best{0.675} & \best{0.756} & \best{0.588} & \best{0.730} & 0.654 & \best{0.730} \\
\bottomrule
\end{tabular*}
\end{table}

\mysection{Further Applications in Medical Imaging Data}{apps}

\begin{figure}[h]
    \centering
    \includegraphics{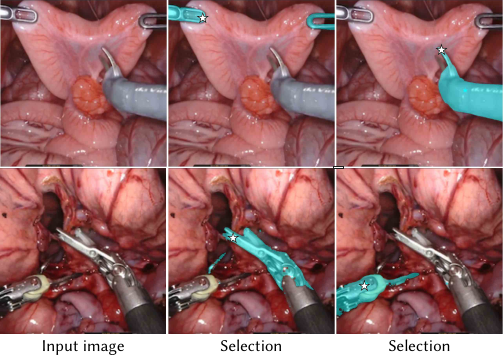}
    \caption{Material selection on medical images. We show that our model generalizes well to extremely out-of-domain examples, such as medical imagery, and that our model is able to output pixel-level accurate masks with simple click operations.}
    \label{fig:supp_medical}
\end{figure}

The SurgVu24~\cite{SurgVu24} challenge released a medical image dataset for classifying and localizing different surgical tools. To demonstrate the practical usefulness of our model beyond image editing tasks, we evaluate whether MAOAM generalizes to out-of-domain images. \cref{fig:supp_medical} shows click-based object selection on surgical imagery. Despite never being trained on medical data, our model produces pixel-accurate masks for surgical tools with simple click interactions, suggesting that the visual grounding learned from our material and object training transfers to novel domains.

\newpage

\end{document}